\newcommand{\sysname}{\textsc{POD-Attention}\xspace}
\newcommand{\sarathipod}{Sarathi+POD\xspace}

\newcommand{\vllm}{vLLM\xspace}
\newcommand{\sarathi}{Sarathi\xspace}

\newcommand{\arxivsummarization}{arXiv-Summarization\xspace}
\newcommand{\arxiv}{arXiv\xspace}

\newcommand{\flashattn}{FlashAttention\xspace}

\newcommand{\faserial}{FA\_Serial\xspace}
\newcommand{\fastreams}{FA\_Streams\xspace}

\newcommand{\flashinfer}{FlashInfer\xspace}
\newcommand{\fiserial}{FI\_Serial\xspace}
\newcommand{\fibatched}{FI\_Batched\xspace}

\newcommand{\hfuse}{FA\_HFuse\xspace}

\newcommand{\sarathiserve}{Sarathi-Serve\xspace}

\newcommand{\llamamha}{Llama-2-7B\xspace}
\newcommand{\llamagqa}{Llama-3-8B\xspace}

\newcommand{\yismall}{Yi-6B\xspace}

\newcommand{\running}[1]{\noindent \textbf{#1:}}
\newcommand{\ignore}[1]{}

\def\checkmark{\tikz\fill[scale=0.4](0,.35) -- (.25,0) -- (1,.7) -- (.25,.15) -- cycle;} 
\documentclass[sigplan,screen]{acmart}
\usepackage[]{hyperref}
\usepackage{amsmath}
\usepackage{subcaption}
\usepackage{graphicx}
\usepackage{xspace}
\usepackage{ifthen}
\usepackage{xcolor}
\usepackage{array}
\usepackage{cleveref}

\usepackage{xparse}
\usepackage{multirow}
\usepackage{booktabs}
\usepackage{arydshln}
\usepackage{enumitem}
\usepackage{algorithm}
\usepackage{algpseudocode}
\makeatletter
\@namedef{ver@lineno.sty}{9999/12/31}
\@namedef{opt@lineno.sty}{}
\makeatother
\usepackage[frozencache,cachedir=.]{minted}
\usepackage{float}
\usepackage{tikz}
\usepackage{paralist}
\usepackage{listings}
\setminted{
linenos=true,
autogobble,
}

\newenvironment{code}
{\minted[escapeinside=@@,xleftmargin=15pt,fontsize=\scriptsize,baselinestretch=0.9]{cuda}}
{\endminted}
\algrenewcommand\algorithmiccomment[1]{\hfill// \textnormal{#1}}

\makeatletter
\renewcommand{\sectionautorefname}{\S\@gobble}
\renewcommand{\subsectionautorefname}{\S\@gobble}
\renewcommand{\subsubsectionautorefname}{\S\@gobble}
\makeatother

\usepackage{ifthen}
\newboolean{publicversion}
\setboolean{publicversion}{true}

\ifthenelse{\boolean{publicversion}}{
    \newcommand{\todo}[1]{}
    \newcommand{\nt}[1]{#1}

    \newcommand{\grumbler}[3]{}
    \newcommand{\ak}[1]{}
    \newcommand{\jm}[1]{}
    \newcommand{\ap}[1]{}
    \newcommand{\rp}[1]{}
    \newcommand{\rr}[1]{}
    \newcommand{\simon}[1]{}
}
{
    \newcommand{\todo}[1]{\textcolor{red}{\bf TODO: #1}}
    \newcommand{\nt}[1]{\textcolor{blue}{#1}}
    \newcommand{\grumbler}[3]{\xspace\textcolor{#3}{\bf #1: #2}}
    \newcommand{\ak}[1]{\grumbler{Aditya}{#1}{violet}}
    \newcommand{\jm}[1]{\grumbler{Jayashree}{#1}{magenta}}
    \newcommand{\ap}[1]{\grumbler{Ashish}{#1}{red}}
    \newcommand{\rp}[1]{\grumbler{Ramya}{#1}{teal}}
    \newcommand{\rr}[1]{\grumbler{Ram}{#1}{teal}}
    \newcommand{\simon}[1]{\grumbler{Simon}{#1}{red}}
}
\AtBeginDocument{%
  }

\copyrightyear{2025}
\acmYear{2025}
\setcopyright{cc}
\setcctype{by}
\acmConference[ASPLOS '25]{Proceedings of the 30th ACM International Conference on Architectural Support for Programming Languages and Operating Systems, Volume 2}{March 30-April 3, 2025}{Rotterdam, Netherlands}
\acmBooktitle{Proceedings of the 30th ACM International Conference on Architectural Support for Programming Languages and Operating Systems, Volume 2 (ASPLOS '25), March 30-April 3, 2025, Rotterdam, Netherlands}\acmDOI{10.1145/3676641.3715996}
\acmISBN{979-8-4007-1079-7/2025/03}

\ignore{
\makeatletter
\gdef\@copyrightpermission{
  \begin{minipage}{0.3\columnwidth}
   \href{https://creativecommons.org/licenses/by/4.0/}{\includegraphics[width=0.90\textwidth]{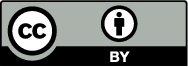}}
  \end{minipage}\hfill
  \begin{minipage}{0.7\columnwidth}
   \href{https://creativecommons.org/licenses/by/4.0/}{This work is licensed under a Creative Commons Attribution International 4.0 License.}
  \end{minipage}
  \vspace{5pt}
}
\makeatother
}

\begin{document}

\title{\sysname: Unlocking Full Prefill-Decode Overlap for Faster LLM Inference}

\author{Aditya K Kamath}
\authornote{Work done as an intern at Microsoft Research India.}
\affiliation{%
  \institution{University of Washington}
  \country{Seattle, USA}}
\author{Ramya Prabhu}
\affiliation{%
  \institution{Microsoft Research}
  \country{Bengaluru, India}}
\author{Jayashree Mohan}
\affiliation{%
  \institution{Microsoft Research}
  \country{Bengaluru, India}}
\author{Simon Peter}
\affiliation{%
  \institution{University of Washington}
  \country{Seattle, USA}}
\author{Ramachandran Ramjee}
\affiliation{%
  \institution{Microsoft Research}
  \country{Bengaluru, India}}
\author{Ashish Panwar}
\affiliation{%
  \institution{Microsoft Research}
  \country{Bengaluru, India}}

\renewcommand{\shortauthors}{Aditya K Kamath et al.}

\begin{abstract}

Each request in LLM inference goes through two phases: compute-bound \textit{prefill} and memory-bandwidth-bound \textit{decode}. To improve GPU utilization, recent systems use hybrid batching that combines the prefill and decode phases of different requests into the same batch. This approach optimizes linear operations but remains inefficient for attention computation because \textit{existing attention kernels specialize execution independently for the prefill and decode phases}.

In this paper, we present \sysname{}
 --- the first GPU kernel that efficiently computes attention for hybrid batches. \sysname aims to maximize the utilization of both compute and memory bandwidth by carefully allocating the GPU's resources such that prefill and decode operations happen concurrently on the same multiprocessor. 
 \sysname  speeds up attention computation by up to $59\%$ (mean 28\%), enabling higher throughput and lower latency LLM inference compared to the use of independently optimized prefill and decode attention kernels.

\end{abstract}




\maketitle

\section{Introduction}

The infrastructure for serving large language models (LLMs) is expanding to meet their growing demands~\cite{ai-infra-spending, ai-infra-spending-2}. Large-scale service providers often depend on expensive high-end GPUs to meet peak demand or latency targets~\cite{patel2023splitwise}. Therefore, optimizing LLM serving systems has become crucial~\cite{orca, vllmsosp, sarathiserve2024, distserve2024, nanoflow2024, loongserve2024, mnemosyne2024}. The overall efficiency of a deployment depends on how well GPU resources are utilized. 

From a resource utilization perspective, LLM inference is a challenging workload because different phases require different resources at different times~\cite{sarathi2023, sarathiserve2024, nanoflow2024, vidur}. The processing of an LLM request begins with a highly parallel (hence, compute-bound) prefill phase which is then followed by a memory-bound decode phase ~\cite{sarathi2023}. 
Serving LLMs efficiently, therefore, requires both high compute and high memory bandwidth. An ideal system would strive to maximize the utilization of both compute and memory. However, doing so is non-trivial because for a given request, the prefill and decode phases occur at different times.

\begin{figure}[t!]
    \centering
    \begin{subfigure}{0.23\textwidth}
    \includegraphics[width=\columnwidth]{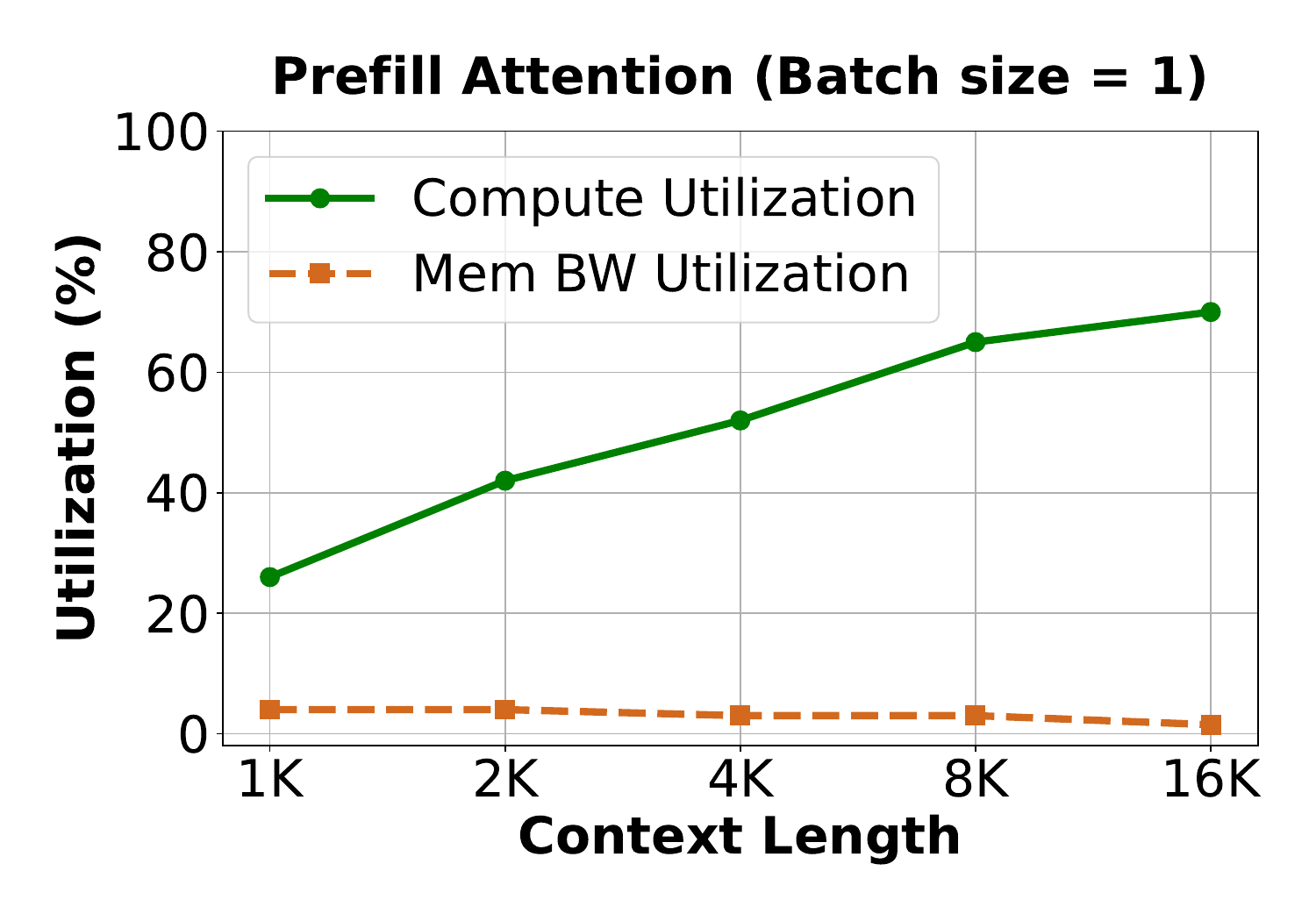}
    \end{subfigure}
    \begin{subfigure}{0.23\textwidth}
    \includegraphics[width=\columnwidth]{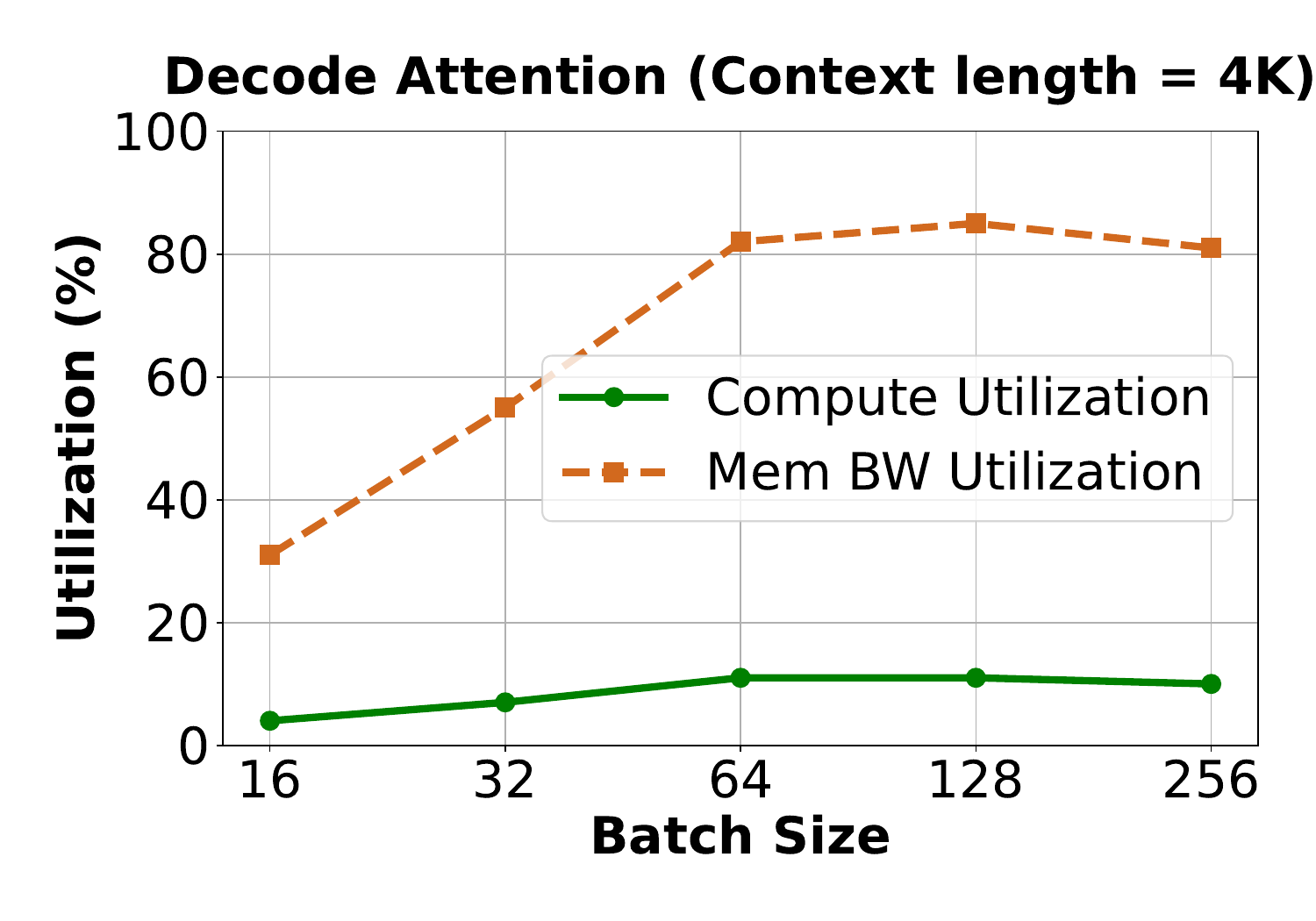}
    \end{subfigure}
    \\
    \begin{subfigure}{0.23\textwidth}
    \includegraphics[width=\columnwidth]{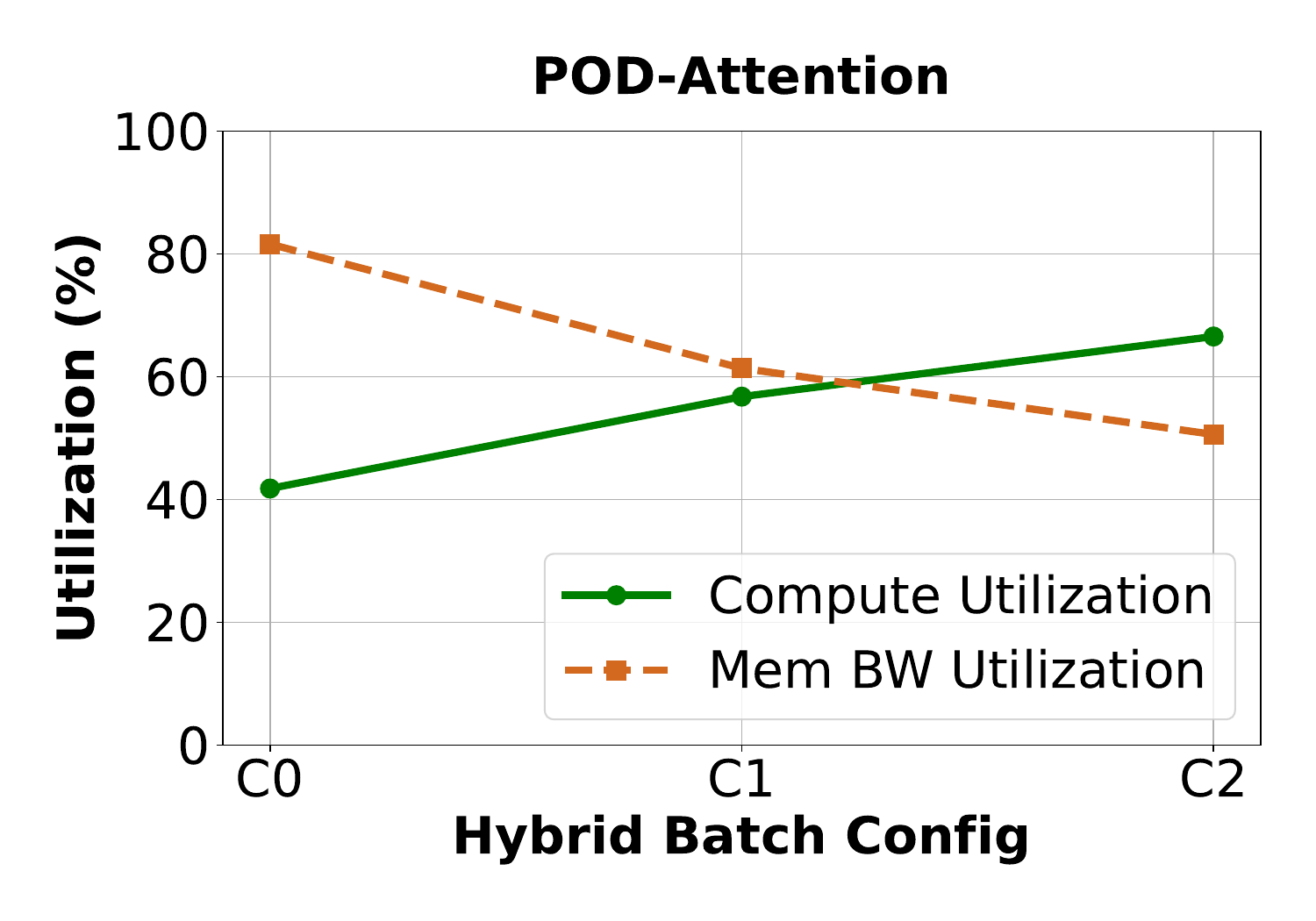}
    \end{subfigure}
    \begin{subfigure}{0.23\textwidth}
    \includegraphics[width=\columnwidth]{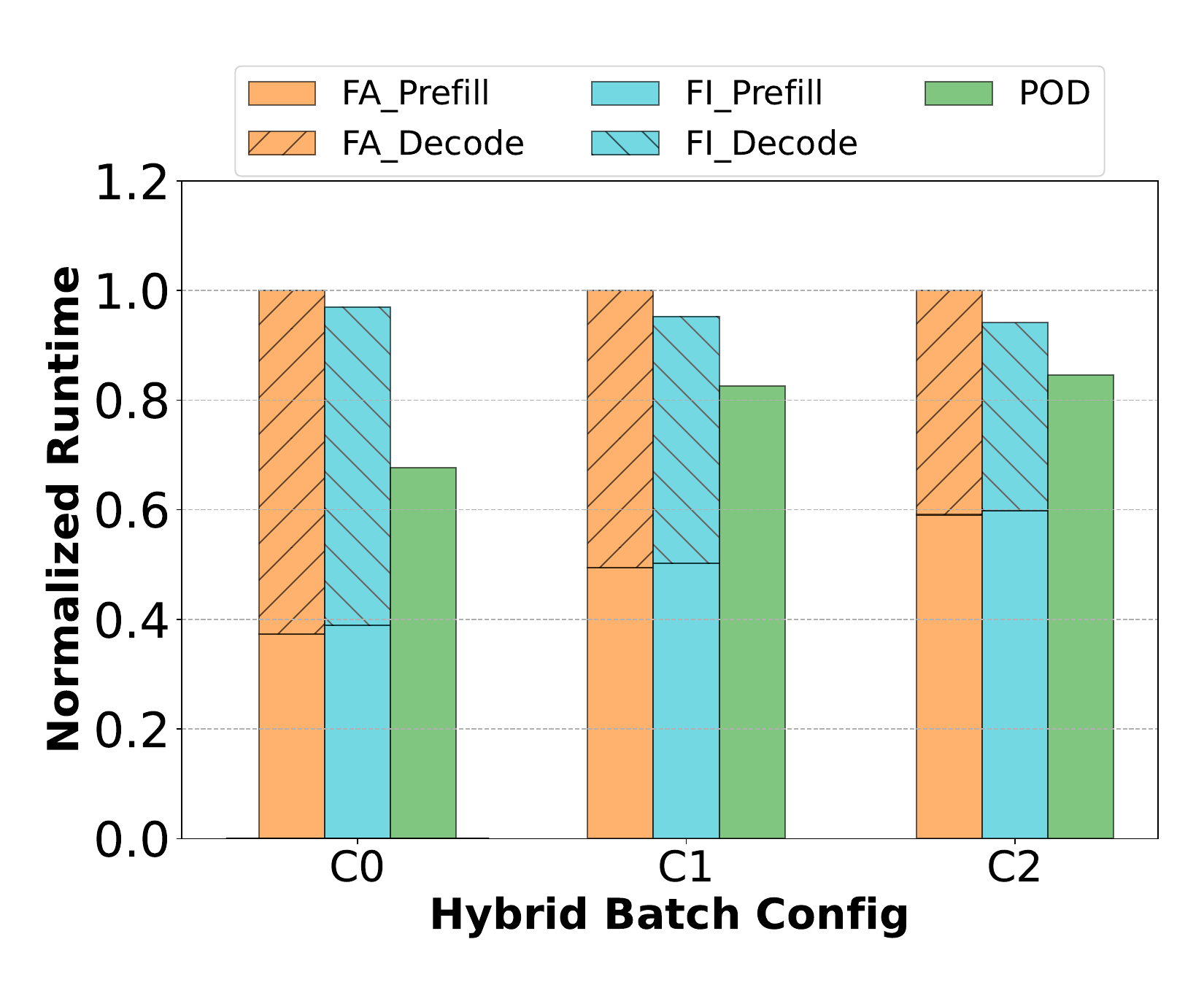}
    \end{subfigure}
    \caption{State-of-the-art attention kernels utilize either compute or memory (FA: FlashAttention, FI: FlashInfer). \sysname utilizes both compute and memory to accelerate attention computation in hybrid batches (see~\autoref{table:banner-configs} for configurations. Model: \llamagqa on 2 A100 GPUs).}
    \label{fig:banner}
\end{figure}

State-of-the-art LLM serving systems deal with this challenge by combining the inputs of prefill and decode phases of different requests into the same batch~\cite{sarathi2023, splitfuse2024, orca} --- a technique we refer to as \textit{hybrid batching}. Hybrid batching avoids the need to fetch model weights from GPU high-bandwidth memory (HBM) separately for prefill and decode tokens. Instead, it allows the GPU to fetch model weights once and use them to compute over both prefill and decode inputs. Hybrid batching also helps reduce tail latency: to limit the runtime of each iteration, the scheduler can  divide long input prompts (prefill inputs) into multiple smaller chunks, then combine ongoing decodes with a new prefill chunk every iteration~\cite{sarathiserve2024, splitfuse2024}. As such, use of hybrid batching is common in various LLM serving systems today~\cite{vllmsosp, nanoflow2024, sarathiserve2024, orca, splitfuse2024}.

While prior work has focused on optimizing the linear operations~\cite{sarathiserve2024,orca,splitfuse2024}, they do not optimize the attention computation of a hybrid batch. This is reasonable for a system that primarily deals with small context lengths since linear operations dominate run time in this setting~\cite{orca, nanoflow2024}. In contrast, as the context length increases, attention computation becomes the primary performance bottleneck~(\autoref{fig:motivation:op-breakdown}).
 
Some recent works have also tried to optimize attention computation~\cite{flashattention,flashdecoding,flashdecoding++,leanattention}, but current solutions address prefill and decode operations separately --- maximizing compute utilization for prefills and bandwidth utilization for decodes, as shown in~\autoref{fig:banner}. In this paper, we show that such an approach is suboptimal as it leaves critical GPU resources underutilized in different parts of computation. For example,~\autoref{fig:banner} illustrates that memory bandwidth utilization of the prefill attention kernel is often below 5\%, while compute utilization of the decode attention kernel is under 10\%. The effect of using independently optimized kernels is particularly noticeable with hybrid batching because prefill and decode kernels execute immediately one after the other, leading to periods of high demand of a resource immediately followed by low utilization of the same resource. 

To improve the efficiency of hybrid batching, we present \sysname{} --- the first GPU kernel, to the best of our knowledge, that efficiently batches the computation of prefill and decode attention. In doing so, we first show~(\autoref{sec:case-study}) that existing techniques do not provide adequate performance in fusing attention computation due to various limitations such as straggler threads, synchronization barriers and lack of guaranteed SM-level co-location of different Cooperative Thread Arrays (CTAs) on GPU Streaming Multiprocessors (SMs). \sysname addresses these issues by fusing the computation in a CTA-parallel manner, introducing SM-aware software-based CTA scheduling within the GPU (\autoref{sec:design}). Building on state-of-the-art FlashAttention kernels~\cite{fagithub}, \sysname significantly accelerates attention computation by utilizing both compute and memory resources as per the requirement of a given batch of requests (see~\autoref{fig:banner}).

\begin{table}[t!]
\scalebox{0.9}{
\begin{tabular}{l|ccc|cc|c}
\multirow{2}{*}{Config.} & \multicolumn{3}{c|}{Prefill} & \multicolumn{2}{c|}{Decode} & Resource \\
 & \multicolumn{1}{c}{BS}& \multicolumn{1}{c}{CS} & \multicolumn{1}{c|}{CL} & \multicolumn{1}{c}{BS} & \multicolumn{1}{c|}{CL} & requirement \\ \toprule
C0 & 1 & 1K & 12K & 80 & 12K & memory-bound \\
C1 & 1 & 12K & 12K & 220 & 12K & balanced \\
C2 & 1 & 16K & 16K & 250 & 12K & compute-bound \\ \bottomrule
\end{tabular}}
\caption{Details of hybrid batches evaluated in~\autoref{fig:banner} (BS: batch size, CS: chunk size, CL: context length).}
\label{table:banner-configs}
\end{table}

We also integrate \sysname in a state-of-the-art LLM inference scheduler \sarathiserve~\cite{sarathiserve2024}. Our experiments show that \sysname computes attention up to $59\%$ faster (mean 28\%) than the prefill and decode attention kernels of \flashattn and \flashinfer. In terms of the end-to-end LLM inference performance, \sysname improves throughput by up to $22\%$ while also reducing crucial latency metrics such as time-to-first-token (TTFT), time-between-tokens (TBT) and the end-to-end request execution latency over \sarathiserve.

\running{Contributions} We make the following contributions:
\begin{compactitem}[\labelitemi]
    \item We highlight that independently optimizing prefill and decode attention kernels is suboptimal for hybrid batching based LLM inference.
    \item We present \sysname{} --- a GPU kernel that computes prefill and decode attention concurrently to utilize both compute and memory bandwidth simultaneously.
    \item We integrate \sysname in \sarathiserve and show that it enables high throughput and low latency LLM inference compared to the use of independently optimized prefill and decode attention kernels.
\end{compactitem}

\section{Background and Motivation}\label{sec:background}
We first discuss why LLM serving systems use hybrid batching and then motivate the need to optimize attention computation. Finally, we provide an overview of GPU execution.

\subsection{Large Language Model (LLM) Inference}

LLMs process user inputs and outputs as tokens, internally represented as vectors. Each request during inference goes through two phases --- prefill  and decode~\cite{orca}. The prefill phase processes the tokens of a user's prompt in parallel and produces the first output token, whose latency is called time-to-first-token (TTFT). Subsequently, the decode phase generates one output token (per-request) per-iteration auto-regressively. The latency taken to generate each output token is called time-between-tokens (TBT). The prefill phase is highly parallel and compute bound while the decode phase is memory bound. Due to the parallel processing of a large number of tokens, the latency of a prefill iteration is generally higher than that of a decode iteration. 

\begin{figure}[t!]
    \centering
    \includegraphics[width=0.9\columnwidth]{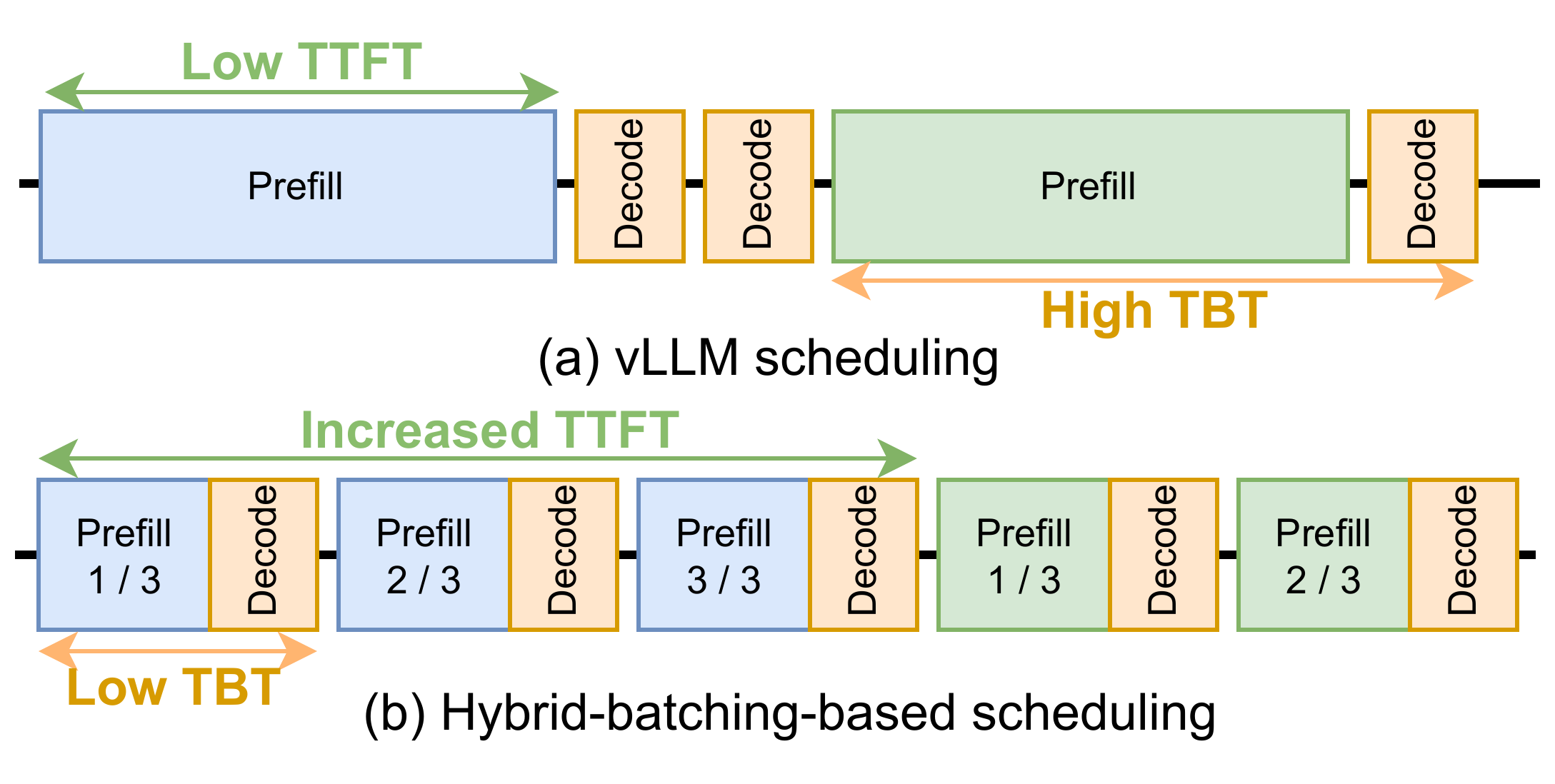}
    \caption{Impact of scheduling strategies on TTFT and TBT.}
    \label{fig:background:scheduling}
\end{figure}

The distinct computational characteristics of prefill and decode operations create a throughput-latency tradeoff in LLM inference~\cite{sarathiserve2024, patel2023splitwise, distserve2024, tetriinfer}, as illustrated in \autoref{fig:background:scheduling}. Since decoding is memory bound, using a large batch size improves throughput. The original \vllm scheduler~\cite{vllmsosp} uses prefill-prioritizing scheduling to maximize the decode batch size (\autoref{fig:background:scheduling}(a)). This approach provides low TTFT, but at the cost of high TBT because a new request's prefill can pause ongoing decodes, causing \textit{generation stalls}~\cite{sarathiserve2024}. High TBT is especially problematic in long-context scenarios, where each generation stall can last several seconds.

The issue of high TBT has been acknowledged in real-world deployments~\cite{vllm:performance-guide-chunking}. \sarathiserve~\cite{sarathiserve2024} proposed \textit{chunked-prefills} coupled with \textit{continuous hybrid batching}~\cite{orca} --- a technique that divides the prefill tokens of a request into multiple smaller chunks and schedules one prefill chunk per-iteration with on-going decodes (\autoref{fig:background:scheduling}(b)). This way, \sarathiserve enables increasing batch size while avoiding generation stalls, improving both performance and user interactivity. Various LLM serving systems have incorporated this technique~\cite{nanoflow2024,sglang2024,trtllmgithub}, including \vllm~\cite{vllm:upstream-chunked-prefill}. 

\nt{In the common case with hybrid batching, an executing batch consists of one prefill chunk of a pre-determined size and multiple decodes (as shown in~\autoref{table:banner-configs}). For example, consider a workload where each request consists of 2K prefill tokens and generates 200 output (decode) tokens. If the prefill chunk size is 1K, a request's prefill completes over two iterations (prefill tokens / chunk size). Upon completion of the prefill phase, it must execute for another 200 iterations --- each iteration corresponding to one output token. In these 200 iterations, 100 requests can complete their prefill phase to join the running batch. This leads to an effective batch size of 101 in the steady state wherein 100 requests execute in their decode phase alongside one prefill chunk of a new request. Executing these hybrid batches requires both high compute (for the prefill chunk) and high memory bandwidth (for the decode requests).}

\begin{figure}[t!]
    \centering
    \includegraphics[width=\columnwidth]{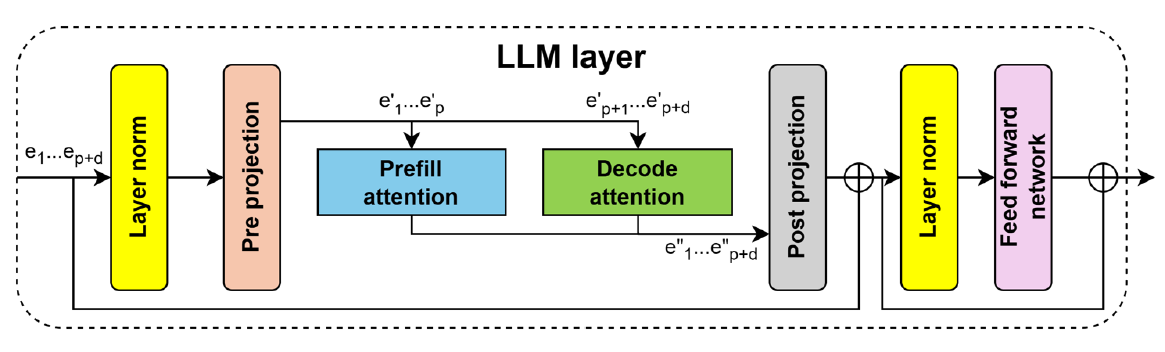}
    \caption{Computation in hybrid batches. Current systems compute prefill inputs ($e_1...e_p$) and decode inputs ($e_{p + 1}...e_{p+d}$) together for linear operations. However, they compute prefill and decode attention separately using specialized kernels.}
    \label{fig:background:hybrid-batching}
\end{figure}

\autoref{fig:background:hybrid-batching} shows how hybrid batching works in practice. Except attention, all other operations are linear i.e., computed element-wise. Linear operations obey the rule f(x + y) = f(x) + f(y) so inputs for a linear operation can be combined, computed upon by the same model weights to reduce memory accesses, and then separated. In contrast, attention is a sequence-level operator that is computed between three representations Q (query, of the current tokens being processed), and K/V (key/value, of all tokens in the sequence seen so far) as:

\[ 
\text{Attention}(Q, K, V) = \text{softmax}\left(\frac{QK^T}{scale}\right) V
\] 

\begin{figure}[t!]
    \centering
    \includegraphics[width=\columnwidth]{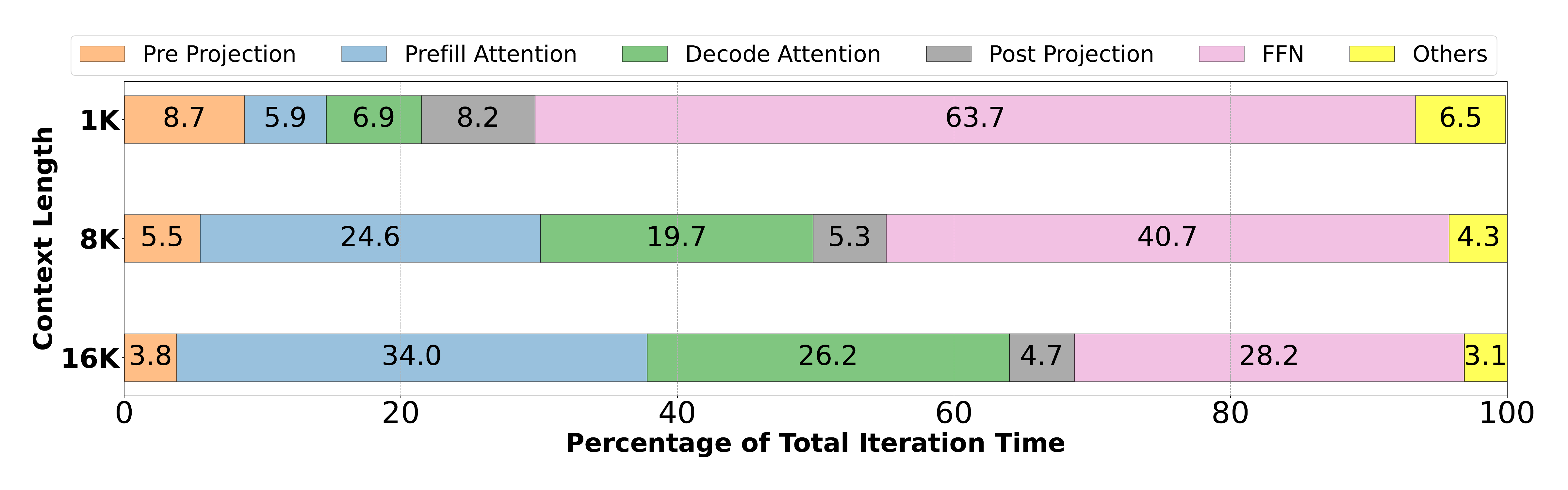}
    \caption{Contribution of different operations in iteration runtime with hybrid batching (model: \llamagqa, batch size: 60, chunk size: 1K). For each context length, we show runtime of iteration that processes the last chunk of a prompt.}
    \label{fig:motivation:op-breakdown}
\end{figure}

The QKV representations are further divided among multiple query heads and K/V heads, each assigned to a group~\cite{ainslie2023gqa}. Attention is computed in parallel for each Q head and K/V head pair. Since resource requirements of prefill and decode attention are different, state-of-the-art libraries such as \flashattn (FA)~\cite{flashattention,flashattention2,flash-attention-3} and \flashinfer (FI)~\cite{flashinfer} provide specialized kernel APIs, optimized separately for each phase. Use of these kernels works well in small context length scenarios where attention computation is a small fraction of the total inference time~\cite{sarathi2023,orca}. 

However, the context length in many real-world LLM applications continues to grow~\cite{mnemosyne2024, loongserve2024}. In such scenarios, attention computation dominates, becoming more than 60\% of the total inference time in many cases as shown in~\autoref{fig:motivation:op-breakdown} (context length 16K). Note that prefill and decode attention are computed immediately one after the other in hybrid batches (see~\autoref{fig:background:hybrid-batching}). Therefore, \textit{when independently optimized attention kernels are used, GPU execution goes through periods of high demand of a resource followed by low utilization of the same resource}. For example, the prefill kernel requires high compute but compute is (mostly) idle when the decode kernel executes. 

We posit that concurrently computing prefill and decode attention can improve performance as it would utilize both compute and memory simultaneously. However, current techniques have several limitations with attention computation. To delve deeper into this, we first explain how GPUs operate and then present a case study of existing methods for executing different operations concurrently on GPUs (\autoref{sec:case-study}).

\subsection{GPU Execution Model}

\begin{figure}[t!]
    \centering
    \includegraphics[width=0.9\linewidth]{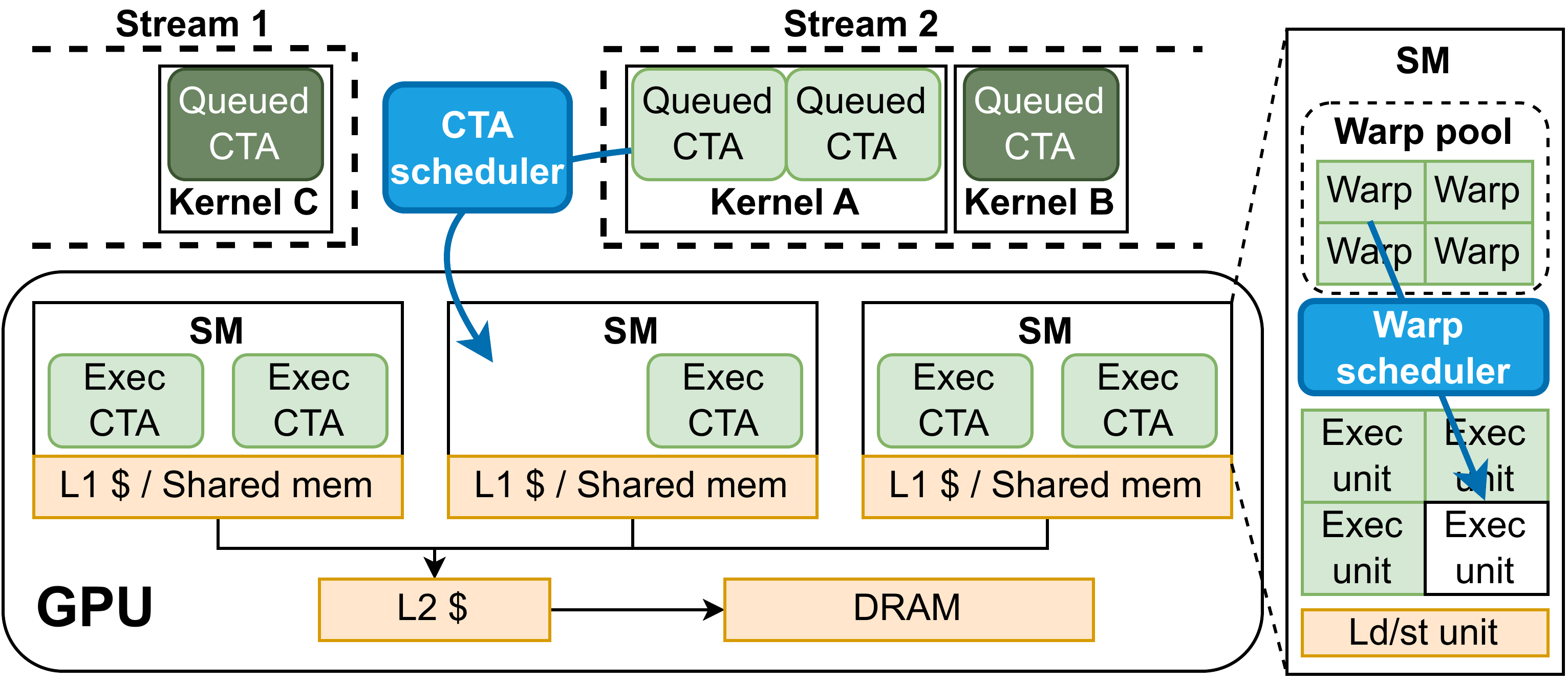}
    \caption{GPU execution model.}
    \label{fig:background:gpu_exec}
\end{figure}

The GPU's hardware is arranged in a hierarchy that supports execution at a scale of hundreds of thousands of parallel threads, depicted in \autoref{fig:background:gpu_exec}~\cite{CUDA_HW}. 
The main processor unit of a GPU is a \textit{Streaming Multiprocessor (SM)}, with modern GPUs containing around a hundred SMs.
Each SM has an L1 cache and \textit{shared memory} along with tensor cores for accelerated general matrix multiplication (GEMM) and execution units for integer/floating point operations.
The shared memory is a user-addressable partition of the L1 cache.
The GPU memory is accessed by SMs through the shared L2 cache.

GPU programming languages expose a hierarchy of threads that mimic the hardware hierarchy. 
The smallest unit of execution is a thread, while a group of 32 threads make up a \textit{warp}, which typically execute concurrently in lockstep.
To maximize throughput, GPU programmers ensure that threads within a warp execute the same code path.
A \textit{Cooperative Thread Array (CTA)}~\cite{PTX_CTA} is a group of warps that share the L1 cache and shared memory.
All warps in a CTA are guaranteed to execute within a single SM.

Users launch GPU \textit{kernels}, or GPU-executed functions, specifying the number of threads in the CTA, the number of CTAs in the kernel, as well as the required shared memory per CTA.
This launch is then queued in a \textit{stream}; operations within a stream are serialized but different streams can execute in parallel in any order. The \textit{CTA scheduler} selects CTAs from streams and assigns them to SMs when sufficient execution resources (e.g., threads, shared memory and registers) are available within the SM.

Central to the GPU's massive throughput is the fast, cycle-level \textit{warp scheduler} baked into the hardware.
Every clock cycle, the warp scheduler dispatches eligible warps for execution; a warp is eligible if its threads aren't stalled (e.g., waiting for memory access).
This allows each SM to context switch at every clock cycle if required, effectively utilizing all its execution resources.
\section{A Case Study on Concurrent Execution}
\label{sec:case-study}

The simplest way to compute prefill and decode attention together is to pass both inputs to an existing attention kernel. Some LLM serving systems prefer this method for computing attention in hybrid batches~\cite{vllm:unifyattentionkernel,sarathi:unifiedattentionkernel}. In~\autoref{sec:eval:attention}, we show that this is counter-productive and slower than serial execution. 

In this section, we focus on GPU methods for concurrent execution e.g., running kernels in parallel or fusing their operations into a single kernel. We quantitatively analyze their performance and highlight key limitations that motivated us to develop a specialized attention kernel.

\subsection{Methods of Concurrent Execution}
\label{sec:coarse-grained-fusion}

Each level of the execution hierarchy in a GPU offers potential for concurrent execution (see \autoref{table:case-study:fusion-methods}). 

\begin{compactenum}[1.]

\item \textbf{Kernel-parallel.} Streams can potentially execute different GPU kernels concurrently. This approach is easy to implement as it only requires submitting existing kernels to different streams; all other approaches require fusing different operations into a single kernel. Unfortunately, streams alone guarantees neither concurrency nor SM-level co-location of different operations~\cite{Elastic_kernel_ASPLOS13, ISPA_TOC23}.

\item \textbf{CTA-parallel.} In this scheme, the CTAs in the kernel are split across operations in a predetermined manner. CTA-parallel enables better load-balancing: when one CTA finishes execution, the GPU scheduler can deploy the next CTA to the SM. However, similar to streams, CTA-parallel does not guarantee SM-level co-location.

\begin{table}[t!]
\centering
\footnotesize
\scalebox{0.9}{
\begin{tabular}{p{2.8cm}|>{\centering\arraybackslash}p{0.4cm}|>{\centering\arraybackslash}p{0.4cm}|c}
\textbf{Execution method} & \textbf{GC} & \textbf{WQ} & \textbf{Notes} \\ \toprule
Streams~\cite{Elastic_kernel_ASPLOS13} & $\times$ & \checkmark & Easiest to implement \\ \hline
CTA & $\times$ & \checkmark & Easy load balancing\\ \hline
Warp (e.g., HFuse~\cite{HFuse_CGO_2022}) & \checkmark & $\times$ & Suffers from straggler problem \\ \hline
Intra-thread~\cite{kernel_fuse, kernel_waver} & \checkmark & $\times$ & Cannot overlap with CTA barriers \\ \hline
SM-aware CTA (Ours) & \checkmark & \checkmark & Minimizes operation interference \\ \bottomrule
\end{tabular}}
\caption{Methods of concurrently executing or fusing different operations along different levels of the GPU execution hierarchy (GC=guarantees op co-location, WQ=reduces wave quantization).} 
\label{table:case-study:fusion-methods}
\end{table}

\begin{table}[t]
\scalebox{0.85}{
\begin{tabular}{p{2cm}|p{7.0cm}}
\textbf{Config.} & \textbf{Description} \\ \toprule
\faserial &  Serial execution with FA kernels \\ \hline
\fastreams & Parallel execution via streams with FA kernels \\ \hline
\hfuse & Horizontally fused FA kernels with HFuse~\cite{HFuse_CGO_2022} \\ \hline
POD (Ours) & Optimized fused computation with our kernel \\ \bottomrule
\end{tabular}}
\caption{Different methods of computing  attention in hybrid batches (FA: \flashattn).}
\label{table:eval:attn-modes}
\end{table}

\item \textbf{Warp-parallel.} Here, warps within each CTA are split across operations, as proposed in horizontal fusion (HFuse \cite{HFuse_CGO_2022}).  This apprach guarantees co-location since all warps in a CTA are guaranteed to reside within the same SM. Unfortunately, warp-parallel fusion suffers from the straggler problem: \nt{an entire CTA must complete execution before it can be replaced by another one; if one or more of its threads or warps lag behind others, the next CTA is delayed. While fusing the prefill and decode attention computation, the fused kernel requires extensive tuning to deal with a large input space of varying batch sizes and context lengths e.g., some hybrid batches may be prefill heavy and others may be decode heavy. Therefore, a fused prefill-decode attention kernel is particularly vulnerable to the straggler effect with warp-parallel fusion.} 

\item \textbf{Intra-thread.} In intra-thread fusion, each thread alternates between executing instructions of different operations~\cite{kernel_fuse, kernel_waver}. In simple cases, this strategy provides the maximum opportunity to overlap different operations. However, attention kernels use CTA-level sync barriers to coordinate fetching data into shared memory. These barriers limit intra-thread fusion as instructions before a barrier cannot be overlapped with those after the barrier. 

\end{compactenum}

\noindent
We now quantitatively analyze the performance of different methods. Unfortunately, no readily available implementation exists for CTA-parallel and intra-thread fusion. Hence, we first analyze kernel-parallel and warp-parallel methods on attention kernels and then investigate other methods.

\subsection{Analysis of Readily Available Methods}
\begin{figure}[t!]
    \centering
    \includegraphics[width=\columnwidth]{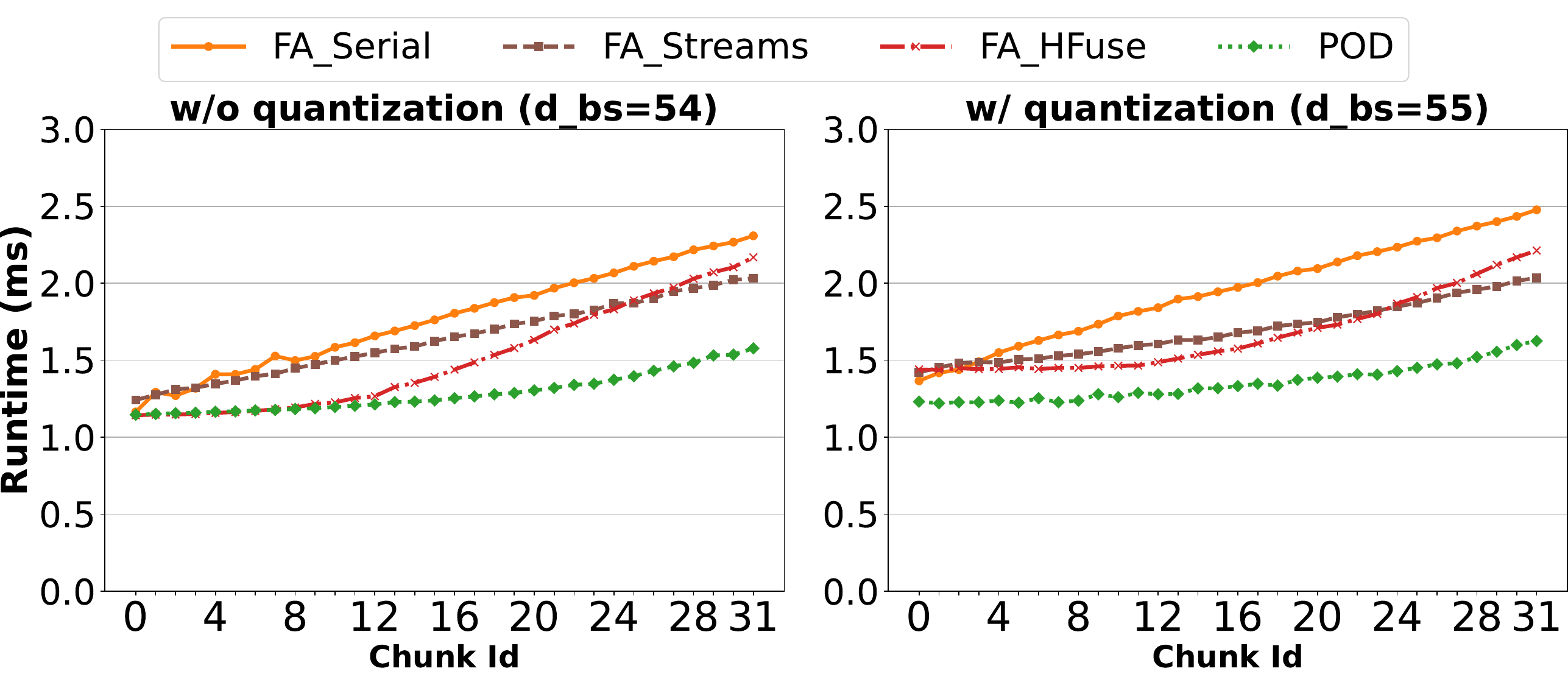}
    \caption{Per layer attention runtime of 32 hybrid batches corresponding to chunked prefills of a request of 16K tokens (chunk size: 512, model: \yismall, d\_bs: decode batch size).}
    \label{fig:motivation:fusing-attn}
\end{figure}

For kernel-parallel execution, shown as \fastreams in~\autoref{fig:motivation:fusing-attn}, we run FA's prefill and decode kernel on two different CUDA streams. For warp-parallel execution (\hfuse), we fuse FA's kernels using the toolchain provided by~\cite{HFuse_CGO_2022}. \autoref{fig:motivation:fusing-attn} compares their performance against serial execution of FA's prefill and decode attention kernels (\faserial).  Our experiment shows the per-layer attention computation time of \yismall for 32 chunks of a 16K prompt (chunk size 512), each co-scheduled with decodes of 16K context length each.

Note that if the number of CTAs in a kernel is not divisible by the number of GPU SMs, some of the SMs in the last wave of scheduling can remain idle --- a phenomenon known as \textit{wave quantization}~\cite{cusync2024,streamk2024}. In the worst case, a marginal increase in work can double the latency of a kernel due to wave quantization. Therefore, to fully understand the benefit of concurrent execution, we evaluate performance with and without wave quantization.  Each decode request uses 4 CTAs in our experiment (one CTA per KV head). Hence a decode batch size of 54 uses 216 CTAs having no wave quantization on our NVIDIA A100 GPU (108 SMs). In contrast, a batch size of 55 uses 220 CTAs leaving 4 quantized CTAs.

\fastreams provides some speed up over \faserial and its gains are higher (up to 20\%) when serial execution suffers from wave quantization. This is because streams run kernels in parallel to fill GPU SMs that would otherwise remain idle. This effect can be seen in~\autoref{fig:motivation:fusing-attn} where \fastreams take roughly the same amount of time for both batch sizes while the time taken by \faserial increases at batch size 55; in particular, decode time increases by more than 25\% in \faserial when batch size goes from 54 to 55 which increases the total attention time of prefill and decode by up to 17\%. \hfuse outperforms \fastreams is some cases but its performance degrades quickly due to straggler effect in the later chunks that are dominated by prefill. This happens because the prefill cost increases with each successive chunk but decode cost is same in all hybrid batches. Overall, \fastreams and \hfuse both perform better than \faserial but still leave significant performance on the table as shown by \sysname which outperforms both methods by a significant margin.

\begin{figure}[t!]
    \centering
    \includegraphics[width=0.9\linewidth]{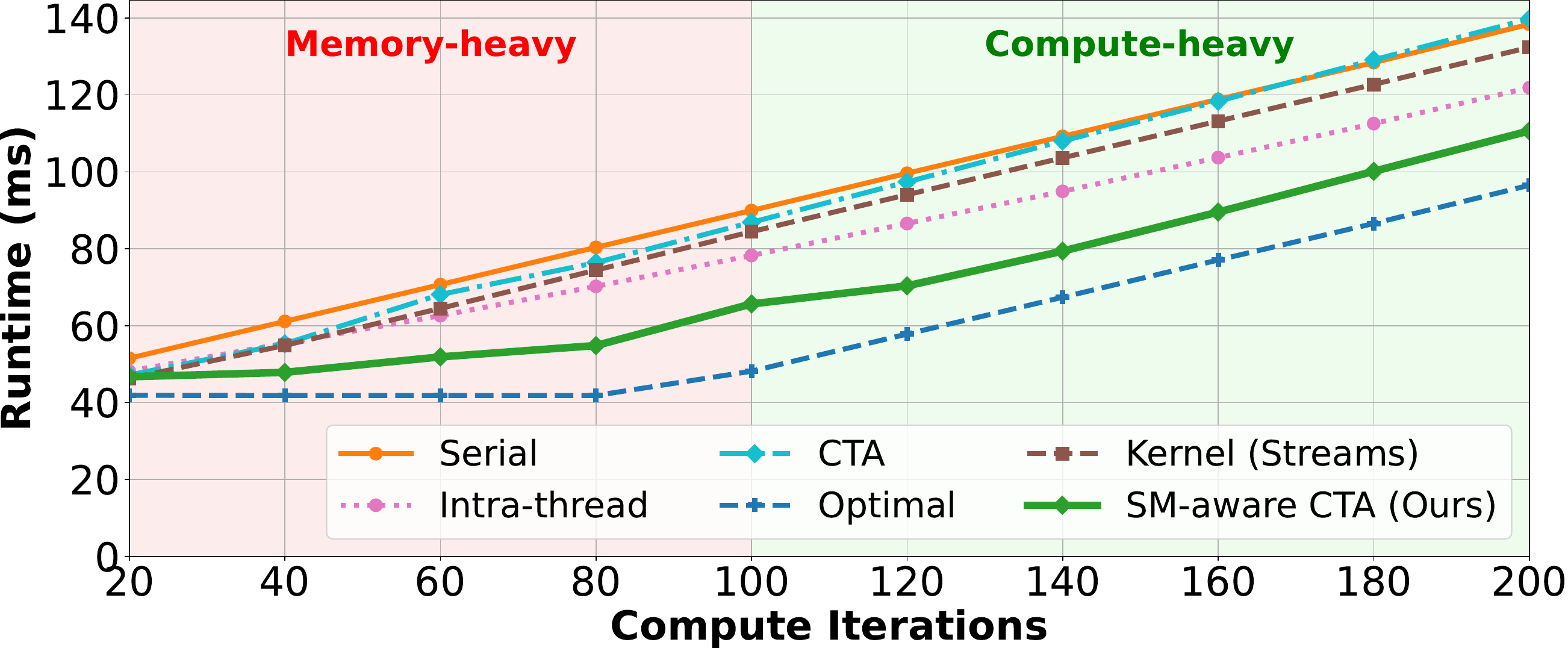}
    \caption{\nt{Fine-grained fusion versus serial computation.}}
    \label{fig:design:parallelism}
\end{figure}

\subsection{Analysis of Other Methods}

For complex kernels, such as attention, efficiently implementing fine-grained fusion schemes is non-trivial and prone to errors. Therefore, we analyze the performance of other fusion methods with a simple micro-benchmark consisting of a compute-bound kernel that repeatedly multiplies array elements with a scalar, and a memory-bound kernel that repeatedly adds three arrays. Each thread executes a barrier after each operation. We vary the number of compute iterations to evaluate performance under varying compositions of compute-bound and memory-bound operations. \autoref{fig:design:parallelism} shows the runtime of different fusion methods applied on these two functions. At 100 compute iterations, both operations consume equal time when executed serially. To the left of this point, memory bound is more dominant. To the right, it is compute bound. \autoref{fig:design:parallelism} also shows the runtime achievable with an ideal oracle (i.e., perfect overlap). 

CTA and kernel-parallel cannot guarantee SM-level co-location of compute-bound and memory-bound operations and hence provides only marginal average improvement of 3\% and 7\% over serial execution. Intra-thread fusion outperforms both serial and CTA-parallel execution, on average by 13\%. However, the benefit of intra-thread fusion is limited due to sync barriers that hinder concurrent execution.

In summary, current methods for concurrently executing heterogeneous operations face several challenges, such as stragglers, barrier-induced delays, and the inability to guarantee SM-level co-location. In the following sections, we demonstrate how a specialized fused kernel, designed to leverage the characteristics of prefill and decode phases, can overcome these challenges.
\section{\sysname}
\label{sec:design}

We introduce \sysname{} --- a single GPU kernel that efficiently computes both prefill and decode attention. Our primary goal is to ensure that each GPU SM computes both operations simultaneously while minimizing resource contention between them. We build our kernel atop FA v2.6.1~\cite{flashattention2}. 

To achieve our goal, we fuse computation along the CTA dimension that helps avoid the pitfalls of finer-grained warp-parallel and intra-thread fusion. In particular, CTA-parallel fusion offers three advantages: 1) it allows different CTAs to start and finish at different times independently of others, 2) ensures that sync barriers do not affect other parts of the computation since the effect of a barrier is limited to within its CTA, and 3) it is easier to program (\autoref{subsec:cta-parallel-implementation}). However, naive CTA-parallel fusion cannot guarantee that prefill and decode will be co-located on GPU SMs. To overcome this limitation, we introduce \textit{software-based SM-aware CTA scheduling wherein each CTA decides whether to compute prefill or decode after it has been dispatched to an SM. }

\begin{figure}[t!]
    \centering
    \includegraphics[width=\linewidth]{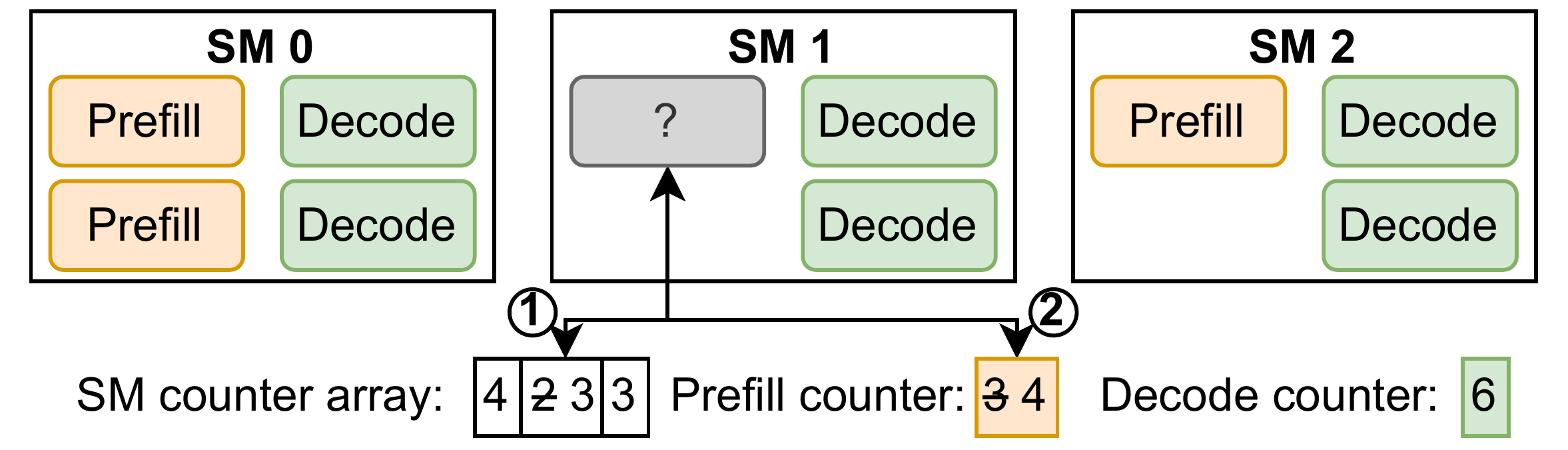}
    \caption{SM-aware CTA scheduling.}
    \label{fig:design:sched}
\end{figure}

\subsection{SM-aware CTA Scheduling} \label{subsec:smaware}

SM-aware CTA scheduling co-locates prefill and decode CTAs through ``runtime operation binding''. Here, a CTA decides whether to perform prefill or decode at runtime, after checking: 1) which SM it got launched on~\cite{smcentric:ICS:2015}, and 2) what other CTAs running on the same SM are doing. This allows the kernel to remain completely agnostic to how the hardware scheduler assigns SMs to CTAs.

To do this, before launching the kernel, we determine how many CTAs are required for prefill and decode independently, and launch the kernel with CTAs matching the sum of both. Each SM has a counter keeping track of the number of CTAs launched on it along with 2 more counters that track the number of prefill and decode CTAs executed on it so far.

\autoref{code:dyn_sched} shows a simple code snippet of SM-aware CTA scheduling. When the hardware scheduler schedules a new CTA on an SM, a leader thread of the CTA (e.g., thread 0) reads the SMID hardware counter~\cite{PTX_SMID} that contains the unique ID of the SM it was launched on (lines 2 - 3). The thread then performs an atomic add operation on the SM counter to obtain a ticket (line 6). This ticket informs the thread as to which operation it should perform i.e., prefill or decode (lines 7 - 8), depending on the scheduling policy. The thread also increments the CTA counter for the operation (line 10). If this exceeds the maximum CTAs for that operation, it switches operations (line 12 - 18). Finally, it writes this information to shared memory so that the other threads in the CTA can begin execution accordingly (lines 20 - 30). We examined two scheduling policies: 50:50 and proportional. In the 50:50 policy, subsequent CTAs on an SM alternate between prefill and decode. In contrast, the proportional policy (line 5) allocates CTAs based on the ratio of prefill and decode CTAs in the current batch.

\begin{figure}[t!]
    \centering
\begin{code}
    if (threadIdx.x == 0) { // Leader thread finds assignment
        int sm_id; // Find which SM this CTA is on
        asm volatile("mov.u32 
        // For this SM, what do we want to run?
        const int ratio = (prefill_ratio + decode_ratio);
        int op, ticket = (atomicAdd(&sm_ctr[sm_id], 1) 
        if(ticket < prefill_ratio) op = PREFILL;
        else op = DECODE;
        // Get the next CTA for operation
        int cta_id = atomicAdd(&cta_assign[op], 1);
        // If the CTA exceeds the max CTA for that op switch ops
        if (op == PREFILL && cta_id >= prefill_ctas) {
            op = DECODE;
            cta_id = atomicAdd(&cta_assign[op], 1);
        } else if (op == DECODE && cta_id >= decode_ctas) {
            op = PREFILL;
            cta_id = atomicAdd(&cta_assign[op], 1);
        }
        // Write the CTA ID and operation to shared memory
        shared_mem[0] = cta_id;
        shared_mem[1] = op;
    }
    __syncthreads(); // Barrier: waits for scheduling to finish
    // Fetch the assigned CTA and operation.
    int cta_id = shared_mem[0];
    const int op = shared_mem[1];
    __syncthreads();
    // Perform the appropriate operation
    if (op == PREFILL) prefill_op(cta_id);
    else decode_op(cta_id)
\end{code}
    \caption{CUDA code for SM-aware CTA scheduling.}
    \label{code:dyn_sched}
\end{figure}

\subsection{Performance Optimizations} \label{subsec:min-interference}
Simply co-locating prefill and decode operations does not yield optimal performance. In this subsection, we introduce various optimizations to maximize the benefit of fusing prefill and decode attention computation. 

\subsubsection{Tile Sizes} \label{subsec:tilesizes}
Data tiling is necessary to make effective use of tensor cores, which provide $\sim$$8\times$ higher throughput than their CUDA core counterpart~\cite{tensorcore}. Tiling also helps improve shared memory usage.
However, the benefit of tiling is not uniform across operations. Decode operates on a single token per request, having a tile length of one across the query sequence length (QSL) dimension.
In Group Query Attention~\cite{ainslie2023gqa}, this length increases to the ratio between query and KV heads, typically 2 -- 8.
Due to this small dimension length, data reuse is insignificant, and performance is limited by memory bandwidth. 

\begin{figure}[t!]
    \centering
    \begin{subfigure}{0.23\textwidth}
    \includegraphics[width=\columnwidth]{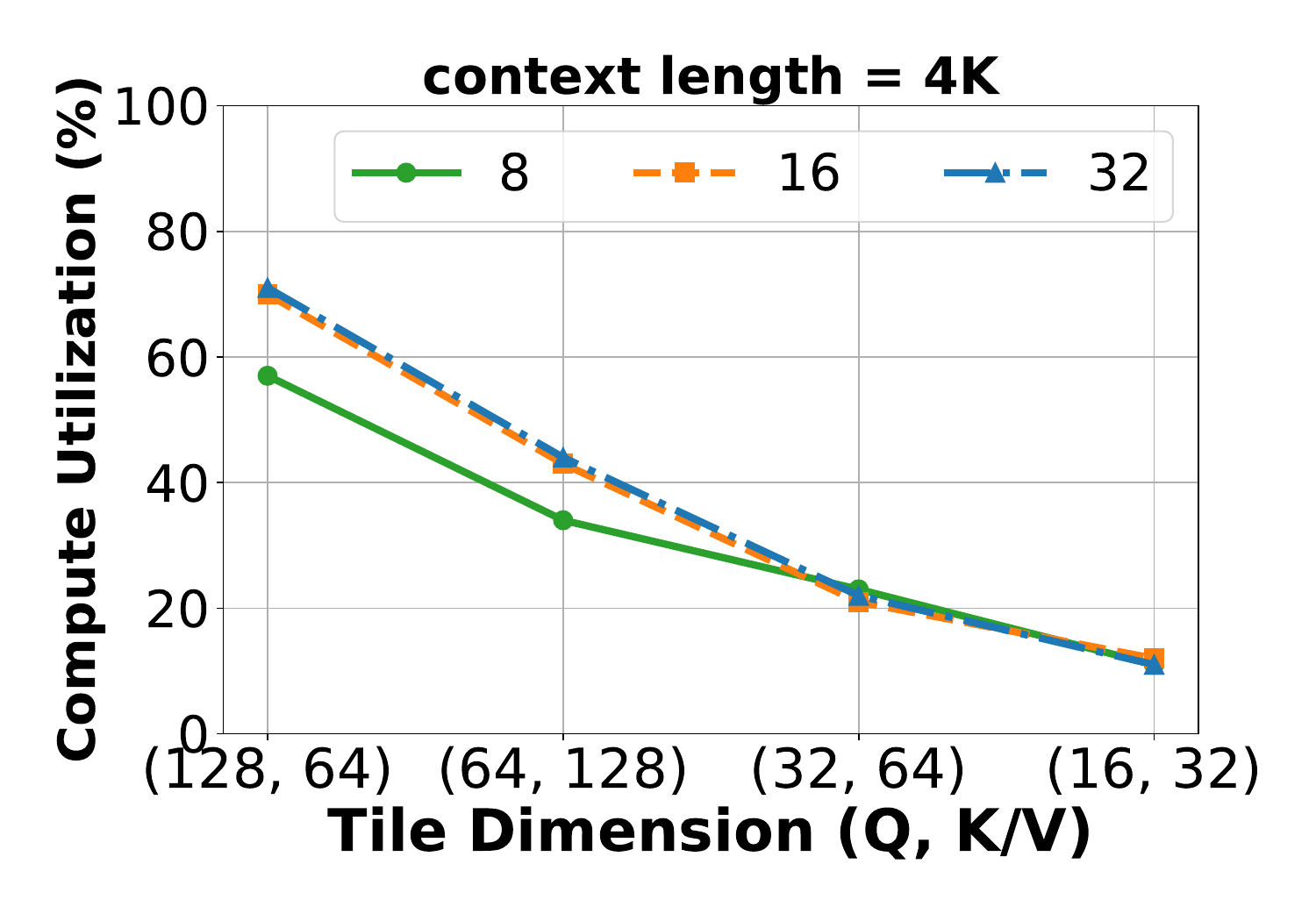}
    \caption{Compute utilization.}
    \label{fig:design:decode_tile_compute}
    \end{subfigure}
    \begin{subfigure}{0.23\textwidth}
    \includegraphics[width=\columnwidth]{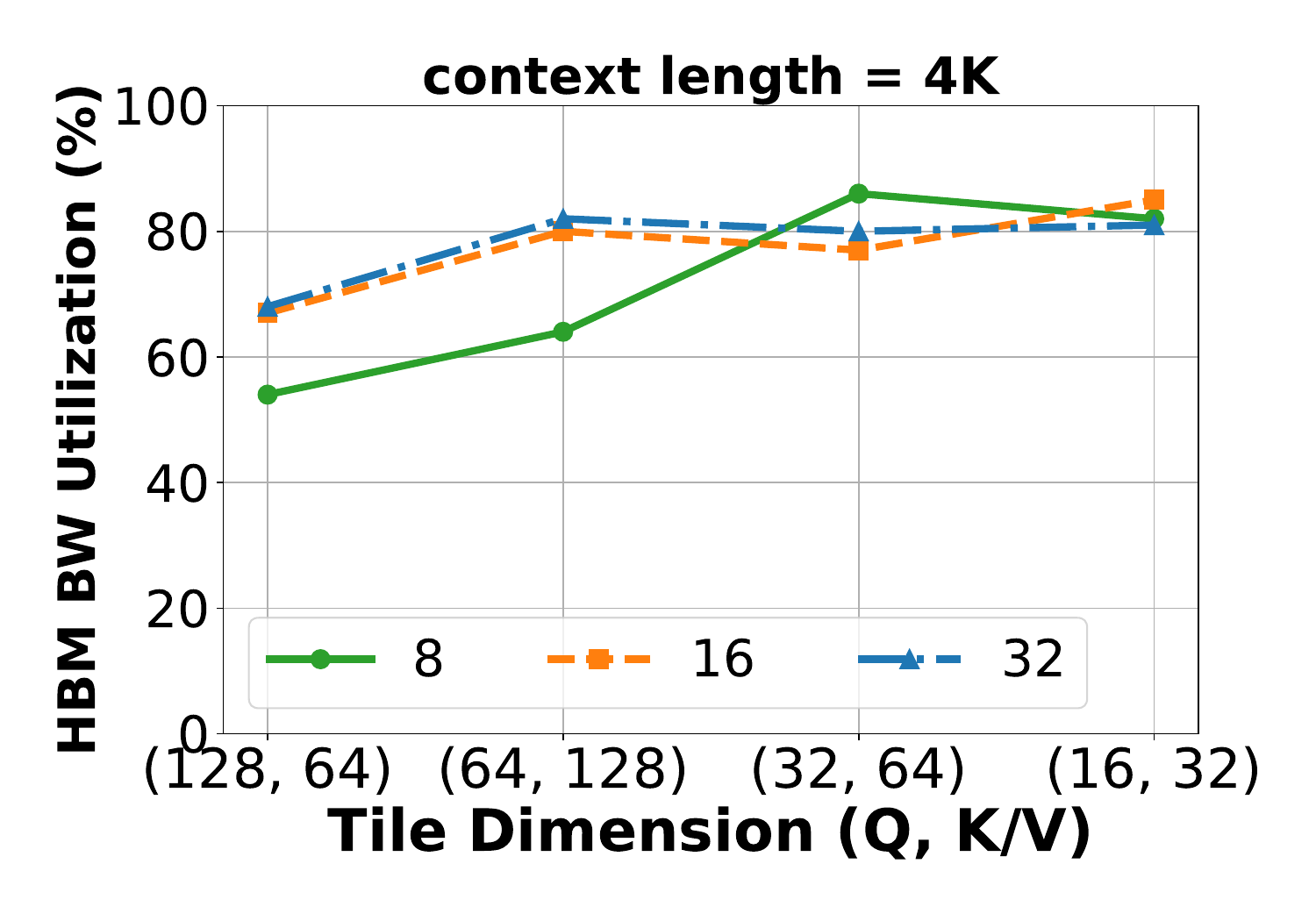}
    \caption{DRAM BW utilization.}
    \label{fig:design:decode_tile_bw}
    \end{subfigure}
    \caption{Impact of decode tile size on compute and HBM BW utilization for batch sizes 8, 16 and 32.}
    \label{fig:design:decode_tile}
\end{figure}

\flashattn uses tile lengths of 64 -- 128 for the QSL dimension. The side-effect of using such large tile sizes is that decodes end up zero padded, causing redundant compute~\cite{flashdecoding++}. For example,~\autoref{fig:design:decode_tile_compute} shows that compute utilization of the decode attention kernel is proportional to tile sizes, reaching up to 70\% at QSL tile dimension of 128, compared to 10\% with tile dimension of 16. However, note that decode attention is memory bound and hence, the primary objective of a decode kernel is to try and saturate memory bandwidth.~\autoref{fig:design:decode_tile_bw} shows that even at a relatively large QSL tile dimension of 64, the decode kernel is able to maximize memory bandwidth utilization. Hence, for a decode-only attention kernel, there is little incentive to reducing tile sizes further.

In contrast, using large tile sizes for decodes is counter-productive in a fused kernel: any redundant compute performed by decodes interferes with co-located prefills since tensor cores are shared between them. If we reduce unnecessary computation, prefill can make better use of the tensor cores. To do so, we use a decode tile length of 16 for QSL, the minimum needed by CUTLASS~\cite{cutlass} for A100 tensor operations. This drops the compute utilization of decodes to $\sim$$10\%$, freeing up tensor cores for prefill.~\autoref{fig:design:decode_tile_bw} shows that reducing tile size has no adverse impact on decode performance at large batch sizes.

\subsubsection{Concurrent CTAs per SM}
The number of CTAs running concurrently on an SM dictates the amount of resources (e.g., shared memory) each CTA can have. More CTAs per SM implies less resources per CTA, but more opportunities for fine-grained scheduling and co-location, i.e., with 2 CTAs per SM we can only co-locate prefills and decodes in a 1:1 ratio, but with 4 CTAs per SM, we can allocate CTAs to prefill and decode in different proportion  depending on batch composition e.g., 3 CTAs to prefill and 1 CTA to decode. In general, prefills benefit from fewer CTAs per SM as it allows each CTA access to more shared memory, enabling use of larger tile sizes. In contrast, decodes do not benefit from larger tile sizes and therefore using more CTAs per SM can be beneficial since it allows fine-grained scheduling. 

To achieve the best of both worlds, \sysname supports two configurations: 2 CTAs per SM for prefill-dominant hybrid batches and 4 CTAs per SM otherwise. Based on the desired configuration, we modify the tile lengths and number of threads used for prefill and decode. We also explored if 8 CTAs per SM can further improve performance and found that it only marginally improves performance in a few cases while under-performing in most cases. \sysname automatically picks the most suitable configuration at runtime.

\subsubsection{Virtual Decode CTAs} 
The amount of shared memory provided to each prefill and decode CTA must be same in the fused kernel. 
However, because decode uses smaller tile sizes, the shared memory requirement of decode is a quarter of the prefill requirement. To avoid over-allocating shared memory to decodes, we divide each decode CTA into virtual CTAs containing a warp of threads. If the original decode CTA has four warps, each virtual CTA contains one warp which uses a quarter of the shared memory of the original CTA. The sum of shared memory used by all the virtual CTAs in each regular CTA is close to the shared memory used by prefill. This way, virtual decode CTAs balance the shared memory used by prefill and decode.

\subsubsection{Limiting Prefill Splits}
FlashAttention parallelizes computation across the query heads and QSL tile dimension. FlashDecoding~\cite{flashdecoding}, designed for decode which has a QSL of one, further splits the computation across the K/V dimension when there is not enough parallelism to fill the SMs of the GPU.  The side-effect of this approach is that different CTAs fetch the same query tensor from memory independently of each other, proportional to the number of splits. Consequently, splitting the computation increases memory bandwidth utilization. While splitting along the key/value dimension is not required for prefills when the input contains enough tokens, chunked-prefills limit the number of tokens processed per-iteration \textit{by design} (to minimize TBT). Therefore, \flashattn also uses the FlashDecoding technique to accelerate the chunked-prefill attention computation. This scheme works well for a prefill-only kernel as increased parallelism can easily offset the cost of extra memory reads.

However, in a fused kernel, using a large number of splits for chunked-prefills can cause memory bandwidth contention between prefill and decode CTAs, potentially negating the benefit of fusion. To balance this trade-off, we limit the number of splits for a chunked-prefill to fill at most two full waves (determined empirically). This allows a chunked-prefill to use more CTAs when required, while ensuring that the number of splits do not get excessive and harm concurrent decodes.

\subsection{Implementing CTA-parallel Fusion} \label{subsec:cta-parallel-implementation}
To fuse the two kernels, we first convert them into generic device functions callable from within GPU code while removing all references to the CUDA-provided CTA ID (i.e., \textit{blockIdx}), instead passing this as a function parameter. We build a wrapper kernel that calls these different functions using a calculated CTA ID. The prefill and decode operations execute as if the supplied CTA ID was their actual ID. This enables flexible remapping of CTA IDs, e.g., CTA 0 of the fused kernel can invoke prefill with CTA ID 0, CTA 1 can call decode with ID 0, CTA 2 can call prefill with ID 1, and so on. The amount of shared memory each CTA gets is fixed at kernel launch time, and prefill and decode operations have different requirements. To manage this, we hand-tune the shared memory usage of both prefill and decode operations to balance their requirements while minimizing performance degradation. We launch our fused kernel with enough shared memory for the maximum needed by either operation. To implement virtual CTAs, we modify the decode function replacing all CTA-level barriers with warp-level barriers. The decode function in the fused kernel is called with the appropriate virtual CTA ID, instead of the assigned CTA ID.

\subsection{Discussion on Alternative Implementations}
\nt{Concurrent execution is a well studied topic in GPU literature~\cite{gemtc:hpdc:2014, Pagoda:ppopp:2017, GPUMultitasking:TPDS:2015, SMK:HPCA:2016}, and our high-level goal of overlapping prefill and decode attention computation can be achieved in multiple ways. One noteworthy strategy is based on persistent threads~\cite{PT:InPar:2012, ISPA_TOC23, Elastic_kernel_ASPLOS13}: in this method, one launches a pre-determined number of CTAs (enough to perfectly fill all the SMs). Persistent threads of these CTAs pull the right type of work as necessary (e.g., prefill or decode tiles). We find that this strategy also alleviates the straggler problem.
However, SM-aware scheduling is still needed to decide what work (prefill or decode) to run on which persistent CTA, critical to guaranteeing operation co-location within an SM. Upon integrating it with SM-aware scheduling, we find that this strategy performs on par with our CTA-parallel fusion. 
}

\ignore{
We prefer CTA-parallel fusion because existing attention kernels make extensive use of CTA tuning. Therefore, CTA-parallel fusion makes it easy to adopt existing kernels e.g., adopting FlashAttention-2 and FlashDecoding kernels (442 and 573 lines of code, respectively) required only 10 and 12 lines of code changes.}

\nt{NVIDIA also provides MPS (multi-process service)~\cite{nvidia:mps} and MIG (multi-instance GPUs)~\cite{nvidia:mig} features to run different applications in parallel on the same GPU. However, because hybrid batching combines prefill and decode operations within a single process by design, MPS and MIG are inapplicable to our use case.}
\section{Evaluation}

\begin{table}[t!]
    \centering
    \scalebox{0.85}{
    \begin{tabular}{l|c|c|c|c}
     Model & GPU & $\#$ Q Heads & $\#$ KV Heads & $\#$ Layers \\ \toprule
    \yismall & 1 A100 & 32 & 4 & 32 \\
    \llamamha & 2 A100s & 32 & 32 & 32 \\
    \llamagqa & 2 A100s & 32 & 8 & 32 \\ \bottomrule
    \end{tabular}}
    \caption{Models and hardware used for evaluation.}
    \label{tab:eval:models}
\end{table}

Our evaluation answers the following questions:
\begin{compactitem}[\labelitemi]
    \item What is the effect of \sysname on attention computation latencies?
    \item How does \sysname affect end-to-end LLM inference performance?
    \item What is the impact of different optimizations and design choices employed in \sysname?
\end{compactitem}

\running{Models and environment} We evaluate \sysname with \yismall (4 KV heads~\cite{yi-6b-200k-hf}), \llamamha (32 KV heads~\cite{llama-2-7b-hf}) and \llamagqa (8 KV heads~\cite{llama-8b-hf}), deploying \yismall on one A100 GPU, and others on two A100 GPUs with tensor parallelism (\autoref{tab:eval:models}). Each model has 32 query heads. Each GPU has 80GB HBM memory.

\running{Workloads and metrics} We evaluate both offline and online inference scenarios. For offline inference, we report the number of requests processed per minute. For online inference, we report TTFT, TBT and request execution latency on two workloads consisting of 2K requests each, and context length ranging from 4K to 32K tokens per-request. One of the workloads is an internal enterprise workload (mean context length of 10.5K tokens, per-request prefill to decode token ratio i.e., P:D in the range of 0 -- 40) and the other is based on \arxivsummarization~\cite{arxiv} (mean context length of 9.5K tokens, P:D ratio of 0-50). On average, the number of decode tokens in \arxiv workload is 42\% higher (470) than the internal workload (331).

\running{Serving system baselines} Our experiments use \sarathiserve~\cite{sarathiserve:github} as the serving framework, which is built atop \vllm~\cite{vLLM:github} . We evaluate two baselines: 1) the original \vllm scheduler~\cite{vllmsosp} that runs prefills and decodes in separate batches, prioritizing prefills over decodes and 2) \sarathiserve~\cite{sarathiserve2024}. Both baselines use FlashAttention kernels (v2.6.1) for attention computation. We integrate \sysname into \sarathiserve to evaluate the  benefits of our optimizations. For simplicity, we refer to \sarathiserve without and with \sysname as \sarathi and \sarathipod. 

\subsection{Evaluating Attention Computation}
\label{sec:eval:attention}

\begin{figure}[t!]
    \centering
    \includegraphics[width=0.99\columnwidth]{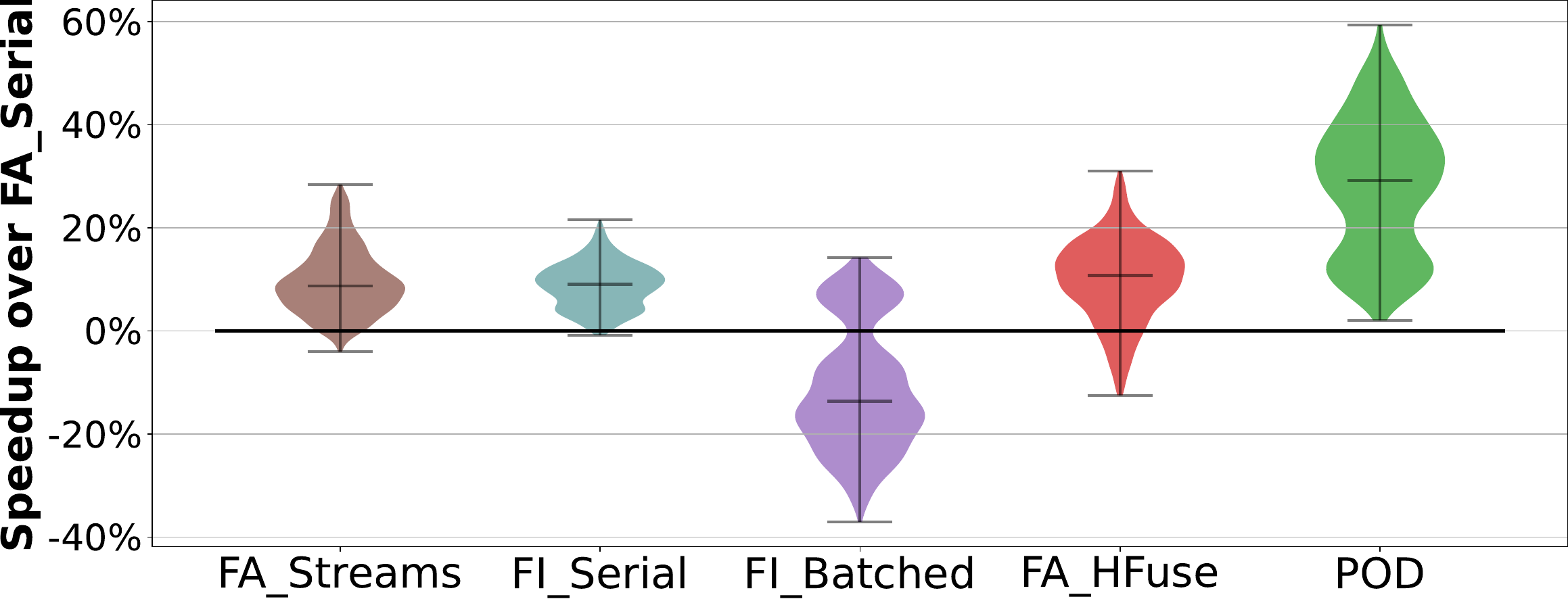}
    \caption{Distribution of speedup in attention computation with different mechanisms compared to \faserial.}
    \label{fig:eval:attention-sweep}
\end{figure}

\autoref{fig:motivation:fusing-attn} illustrates a specific instance where \sysname accelerates attention computation, outperforming the next best alternative by up to 29\%. To demonstrate the broad applicability of \sysname, we conducted a comprehensive sweep across over a thousand hybrid batches on our models. In these experiments, we varied the context length from 4K to 20K and the prefill chunk size from 512 to 2K. We focused on scenarios where prefill and decode attention account for at least 20\% of the serial runtime, as other cases offer limited potential for optimization through operation fusion.

In addition to \flashattn kernels, we also compare the runtime of \flashinfer (FI) v0.2.0 kernels~\cite{flashinfer} in two configurations: \fiserial and \fibatched. \fibatched computes prefill and decode attention using the prefill kernel of \flashinfer. We compare against \fibatched for two reasons: 1) this strategy is the easiest way to compute prefill and decode attention together, and 2) some systems prefer this method e.g., \sarathi used \fibatched in its default attention back-end~\cite{sarathi:unifiedattentionkernel}, and a similar feature is requested in \vllm~\cite{vllm:unifyattentionkernel}. However, we show that this strategy is inefficient e.g., when \fibatched uses a prefill-optimized kernel, it leads to redundant compute in decode computation due to use of larger tile sizes (\autoref{subsec:tilesizes}). This redundant computation interferes with co-running prefill. Similar interference occurs on memory-bandwidth if \fibatched uses a decode-optimized kernel.

\autoref{fig:eval:attention-sweep} shows the relative speedup for different mechanisms compared to \faserial. \fastreams provides limited speedup as it cannot guarantee SM-level overlap of operations. In rare cases, we find that the overhead of stream synchronization can also negate its benefits. 
\fiserial has better optimized decode kernels giving it a modest improvement over \faserial, but it does not overlap the operations.
\fibatched improves performance at low context lengths, but degrades at higher lengths by up to 40\% due to redundant computation for decodes. \hfuse is the strongest baseline as it guarantees operation overlap, improving median performance by 11\%. However, \hfuse is susceptible to the straggler effect due to which it is slower by up to 13\% compared to \faserial. The straggler effect can also be seen in~\autoref{fig:motivation:fusing-attn} towards the later chunks where prefill is more dominant, making it hard to achieve perfect utilization.

\begin{figure}[t!]
    \centering
    \includegraphics[trim={0 0 0 20}, clip, width=0.95\columnwidth]{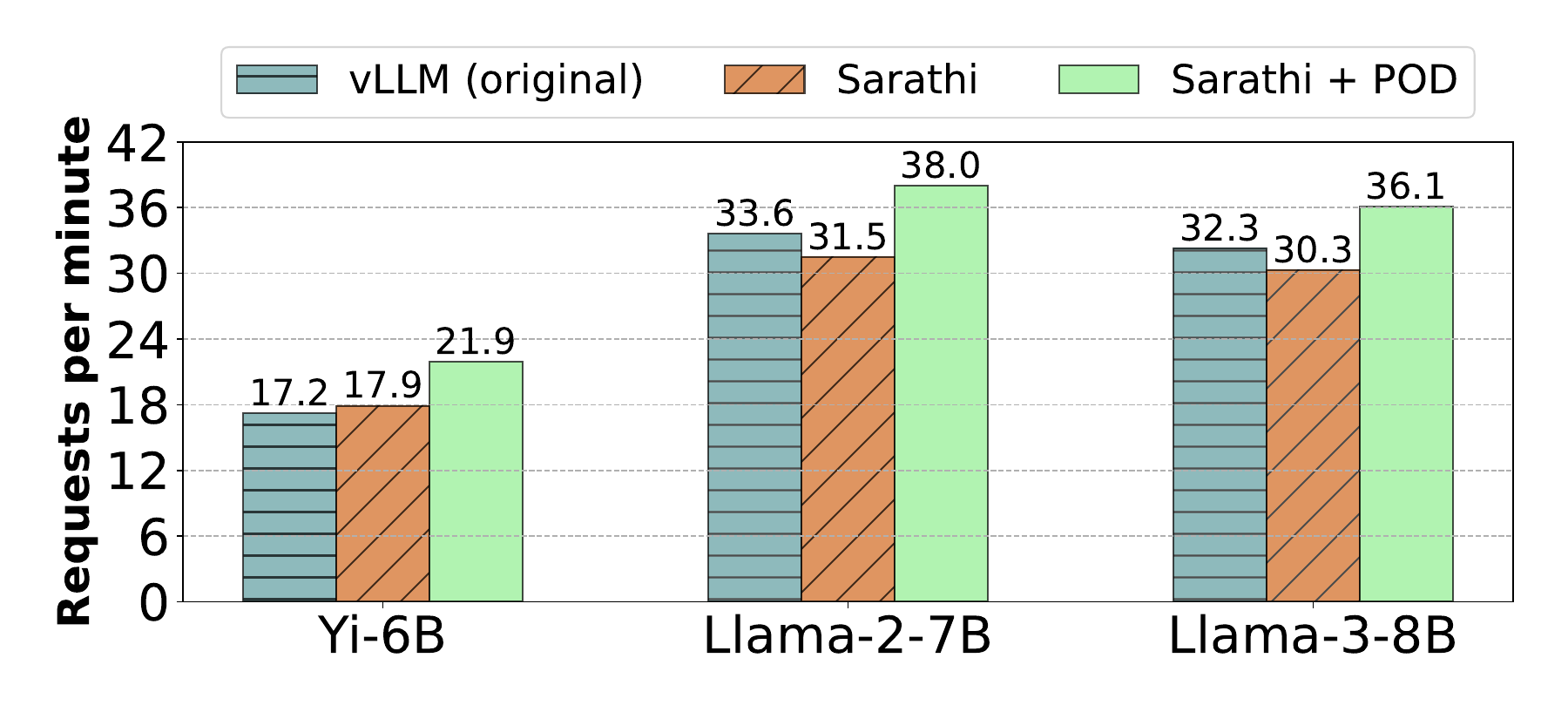}
    \caption{Serving throughput in offline inference.}
    \label{fig:eval:throughput:offline}
\end{figure}

\begin{table*}[t]
\scalebox{0.9}{
\begin{tabular}{l|r|cc|cc|cc|cc}
\multirow{2}{*}{\textbf{QPS}} & \multirow{2}{*}{\textbf{System}} & \multicolumn{2}{c|}{\textbf{TTFT}} & \multicolumn{2}{c|}{\textbf{TBT}} & \multicolumn{2}{c|}{\textbf{Request Latency}} & \multicolumn{2}{c}{\textbf{$\%$ Requests with Stalls}} \\
 &  & \multicolumn{1}{c}{P50} & \multicolumn{1}{c|}{P99} & \multicolumn{1}{c}{P50} & \multicolumn{1}{c|}{P99} & \multicolumn{1}{c}{P50} & \multicolumn{1}{c|}{P99} & \multicolumn{1}{c}{200ms} & \multicolumn{1}{c}{500ms} \\ \toprule
\multirow{3}{*}{1.1} & \vllm (original) & 0.67 & 10.11 & 0.04 & 1.13 & 25.05 & 91.01 &  99.95&  97.8\\
 & \sarathi & 2.2 & 12.58 & 0.10 & 0.15 & 26.83 & 92.24 &  2.05&  0\\
 & \sarathipod  & 1.9 & 12.26 & 0.10 & 0.14 & 24.70 & 79.04 &  3.17&  0\\ \hdashline
 \multirow{3}{*}{1.2} & \vllm (original) & 0.94 & 12.70 & 0.07 & 1.76 & 42.73 & 151.8 &  99.95&  99.6\\
 & \sarathi & 25.44 & 57.83 & 0.12 & 0.16 & 67.12 & 140.5 &  5.07&  2.63\\
 & \sarathipod & 7.49 & 23.78 & 0.11 & 0.15 & 38.69 & 106.8 &  2.29&  0\\ \bottomrule
\end{tabular}}
\caption{Internal workload. Latency numbers in seconds.}
\label{table:eval:online:internal}
\end{table*}

\begin{table*}[t]
\scalebox{0.9}{
\begin{tabular}{l|r|cc|cc|cc|cc}
\multirow{2}{*}{\textbf{QPS}} & \multirow{2}{*}{\textbf{System}} & \multicolumn{2}{c|}{\textbf{TTFT}} & \multicolumn{2}{c|}{\textbf{TBT}} & \multicolumn{2}{c|}{\textbf{Request Latency}} & \multicolumn{2}{c}{\textbf{$\%$ Requests with Stalls}} \\
 &  & \multicolumn{1}{c}{P50} & \multicolumn{1}{c|}{P99} & \multicolumn{1}{c}{P50} & \multicolumn{1}{c|}{P99} & \multicolumn{1}{c}{P50} & \multicolumn{1}{c|}{P99} & \multicolumn{1}{c}{200ms} & \multicolumn{1}{c}{500ms} \\ \toprule
 \multirow{3}{*}{0.85} & vLLM (original) & 0.55& 6.26& 0.03& 0.82& 20.53& 234.93&  99.9&  97.8\\
 & Sarathi & 2.68& 14.89& 0.08& 0.13& 27.87& 281.07&  4.15&  2.05\\
 & \sarathipod & 1.85& 12.71& 0.08& 0.11& 24.31& 255.75&  1.85&  1.61\\ \hdashline
 \multirow{3}{*}{0.95} & vLLM & 0.71& 8.25& 0.06& 1.36& 36.86& 401.2&  99.9&  99.45\\
 & Sarathi & 46.22& 144.2& 0.1& 0.14& 90.12& 417.6&  4.44&  1.9\\
 & \sarathipod & 11.74& 27.38& 0.09& 0.12& 40.6& 333.0&  2.2&  2.1\\ \bottomrule
\end{tabular}}
\caption{\arxiv-based workload. Latency numbers in seconds.}
\label{table:eval:online:arxiv}
\end{table*}

\sysname reaches a peak speedup of 59\%, and a mean of 28\% --- higher than all alternatives.  We found that in 25\% of cases, it also reaches within 10\% of the theoretical peak speedup, signifying near-perfect overlap. Furthermore, unlike other alternatives, \sysname never under-performs serial execution. These results underline the importance of a specialized attention kernel for hybrid-batching-based LLM inference. 

\nt{Additionally, we profiled the energy consumption of the attention kernels and observed that \sysname reduces energy consumption by up to 35\% over \faserial (mean 20.5\%). These savings are largely proportional to the reduction in runtime, showing that prefill-decode overlap not only improves performance but also reduces energy consumption.}

\subsection{Evaluating Throughput in Offline Inference}
\label{sec:eval:offline}

For evaluating offline inference scenarios, we run long context requests of 16K tokens each. We use chunk size 512 for \yismall, and 1K for both \llamamha and \llamagqa, chosen in a way that chunking a prompt does not reduce the performance of linear operations (as recommended by \sarathi~\cite{sarathi2023, sarathiserve2024}). We run 1K total requests for \yismall, and 2K requests each for \llamamha and \llamagqa such that the total runtime of a single configuration is about one hour. The number of output tokens per-request is set to 2K for \yismall, 1K for \llamagqa and 256 for \llamamha; we study the effect of varying prefill to decode token ratio (P:D ratio) in~\autoref{sec:eval:ablation:varyingpd}.

\autoref{fig:eval:throughput:offline} shows that \sarathipod delivers the best throughput: $22\%$, $20\%$ and $19\%$ higher than \sarathi, and $27\%$, $13\%$ and $12\%$ higher than \vllm, for the three models. It is  worth highlighting that chunked-prefills and hybrid batching involves a tradeoff. Chunking a prompt increases attention computation time due to repeated KV cache loads: computing attention of a prefill chunk requires reading KV cache of all prior chunks~\cite{distserve2024}. At the same time, fusing decode tokens with prefills helps execute linear operations more efficiently: model weights need not be read separately for prefills and decodes. Therefore, the relative performance of \vllm and \sarathi can vary depending on workload, model configuration and chunk size. In our experiments, \sarathi improves throughput slightly over \vllm for \yismall but under-performs it for \llamamha and \llamagqa. \sarathipod fuses prefills and decodes in all operations to improve GPU resource utilization, thereby outperforms both baselines.

\subsection{Evaluating Latency in Online Inference}
\label{sec:eval:online}

We evaluate \llamagqa on the internal and \arxiv-based workloads near the serving capacity of the system: the maximum load a system can handle while avoiding high queuing delays~\cite{sarathiserve2024}. We evaluate 2048 requests in each workload by varying the input load based on Poisson distribution. For \sarathi and \sarathipod, we use chunk size of 1024 for the \arxiv-based workload, and 1536 for the internal workload which is more prefill-heavy. We discuss performance on important LLM-specific latency metrics of TTFT, TBT, and end-to-end request execution latency.

Note that there is an inherent trade-off between these metrics~\cite{sarathiserve2024} and  optimizing for one metric can severely compromise the others. For example, as will see below, \vllm  prioritizes prefills and thus achieves low TTFT but sacrifices TBT, resulting in 95+\% of user requests experiencing one or more stalls during decode generation. On the other hand, \sarathi reduces the stalls to a small \% of user requests but significantly increases TTFT compared to \vllm. 

\subsubsection{TTFT} \vllm provides the lowest TTFT as it schedules a prefill on the first available opportunity. In comparison, \sarathi increases TTFT because the ongoing decodes interfere with prefills. TTFT in \sarathi further increases with the load, particularly due to higher queuing delays, e.g., the median TTFT goes to 25.4 and 46.2 seconds for the internal and \arxiv-based workloads, compared to 0.94 and 0.71 seconds of \vllm. \sarathipod significantly reduces TTFT  over \sarathi, bringing the median TTFT down to 7.5 and 11.74 seconds at higher load. \sarathipod also reduces the P99 TTFT by up to $4.3\times$ over \sarathi.

\subsubsection{TBT and Stalls} \vllm induces generation stalls by pausing on-going decodes whenever a new prefill is scheduled, resulting in poor interactivity with the LLM service. These generation stalls are reflected as high tail TBT latency, e.g., the P99 TBT of \vllm reaches up to 1.76 seconds (internal workload) and 1.36 seconds (\arxiv-based workload). In the worst-case, we observe that the highest TBT latency reaches up to 8 seconds in \vllm when it computes multiple prefills consecutively. In comparison, \sarathi ensures that ongoing decodes do not get affected by a new prefill. Therefore, \sarathi provides significantly lower tail TBT latency compared to \vllm e.g., the P99 TBT of \sarathi is at most 0.16 seconds ($10\times$ lower than \vllm). \sarathipod further minimizes tail TBT over \sarathi by 10 -- 20\%. Crucially, since a single response results in a large number of decodes, {\it high TBT tail latency affects nearly all requests in \vllm }, signifying poor interactive experience for almost all users. Even if the TBT SLO is raised to 500ms, more than 97\% of the total requests experience at least one stall in \vllm. In contrast, very few requests (<5\%) observe a stall in \sarathi, which \sarathipod further reduces in most cases.

\begin{table}[t!]
\scalebox{0.95}{
\begin{tabular}{l|c|c|c|c}
\multirow{2}{*}{\begin{tabular}[c]{@{}l@{}}\textbf{Latency}\\ \textbf{Metric}\end{tabular}} & \multirow{2}{*}{\begin{tabular}[c]{@{}c@{}}\textbf{\vllm}\\ (original)\end{tabular}} & \multicolumn{3}{c}{\textbf{\sarathipod}} \\
 &  & 1024 & 1536 & 2048 \\ \toprule
TTFT (P50) & 0.67 & 6.29 & 1.9 & 1.59 \\
TTFT (P99) & 10.11 & 18.99 & 12.26 & 12.40 \\ \hdashline
TBT (P50) & 0.04 & 0.08 & 0.10 & 0.08 \\
TBT (P99) & 1.13 & 0.11 & 0.14 & 0.18 \\ \bottomrule
\end{tabular}}
\caption{TTFT and TBT of \sarathipod with different chunk sizes versus \vllm (internal workload, QPS 1.1).}
\label{table:eval:online:chunk-size}
\end{table}

\subsubsection{End-to-end Request Latency} Request latency can be used to approximate system throughput in online inference. \sarathi reduces P99 request latency over \vllm by 8\% for the internal workload at QPS 1.2, but increase it by up to 24\% over \vllm for the \arxiv-based workload (QPS 0.85). \sarathipod is not only better than \sarathi in all cases, but also outperforms \vllm in many cases e.g., it reduces the P99 request execution latency by up to 42\% over \vllm for the internal workload (106.8 seconds vs 151.8 seconds at QPS 1.2) and by up to 17\% for the \arxiv-based workload (333 seconds vs 401.2 seconds at QPS 0.95).

These results demonstrate that \sarathi enhances interactivity by reducing tail TBT and minimizing stalls, albeit with increased TTFT and some throughput reduction compared to \vllm. \sysname optimizes \sarathi’s performance across all metrics, effectively balancing the throughput-latency tradeoff.~\autoref{table:eval:online:chunk-size} shows that the chunk size in \sarathipod can be tuned further to navigate the TTFT and TBT trade-off, e.g., using a larger chunk size of 2K tokens lowers the median TTFT from 6.3 seconds to 1.6 seconds at the cost of higher TBT (P99 0.18 seconds vs 0.11 seconds).

\subsection{Sensitivity Studies}

\begin{figure}[t!]
    \centering
    \ignore{
    \begin{subfigure}{0.23\textwidth}
        \includegraphics[width=\columnwidth]{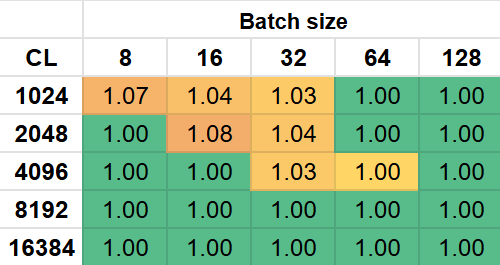}
        \caption{MHA: 2 CTAs per SM.}
        \label{fig:eval:sens_cta_2}
    \end{subfigure}
    \hfill
    \begin{subfigure}{0.23\textwidth}
        \includegraphics[width=\columnwidth]{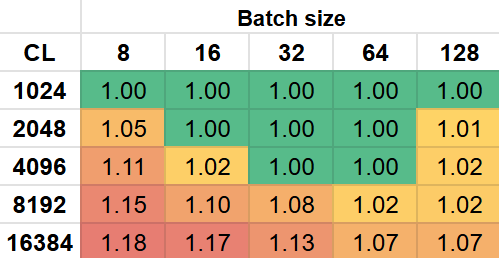}
        \caption{MHA: 4 CTAs per SM.}
        \label{fig:eval:sens_cta_4}
    \end{subfigure}
    }
    \begin{subfigure}{0.23\textwidth}
        \includegraphics[width=\columnwidth]{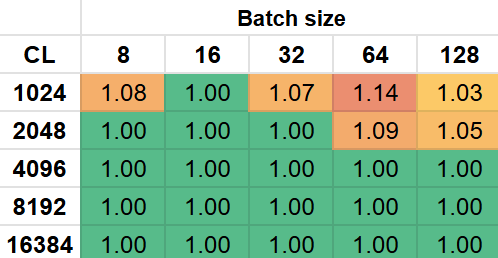}
        \caption{2 CTAs per SM.}
        \label{fig:eval:sens_cta_2}
    \end{subfigure}
    \hfill
    \begin{subfigure}{0.23\textwidth}
        \includegraphics[width=\columnwidth]{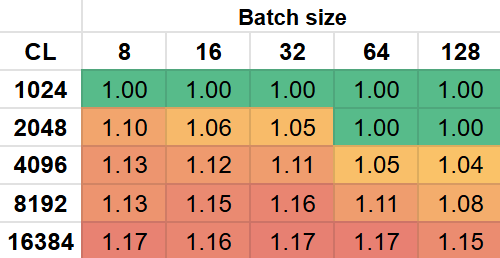}
        \caption{4 CTAs per SM.}
        \label{fig:eval:sens_cta_4}
    \end{subfigure}
    \caption{\sysname with varying CTA configs.}
    \label{fig:eval:sens_cta}
\end{figure}

\subsubsection{CTAs per SM} 
\autoref{fig:eval:sens_cta} shows the performance of \sysname with different numbers of CTAs running concurrently on an SM, varying batch sizes (horizontally) and context lengths (vertically) for \llamagqa.
For each (context length, batch size) data point, we normalize the runtime to the best among the two configurations. In general, for long contexts where prefill cost dominates, 2 CTAs per SM performs better as it allows for larger tile sizes. As the context length decreases, the decode cost starts demonating and hence 4 CTAs per SM starts performing better: more CTAs per SM allows packing more decodes with fewer prefills, e.g., 1 prefill CTA and 3 decode CTAs.

\subsubsection{Scheduling Policy} 
\begin{figure}[t!]
    \centering
    \includegraphics[width=0.9\columnwidth]{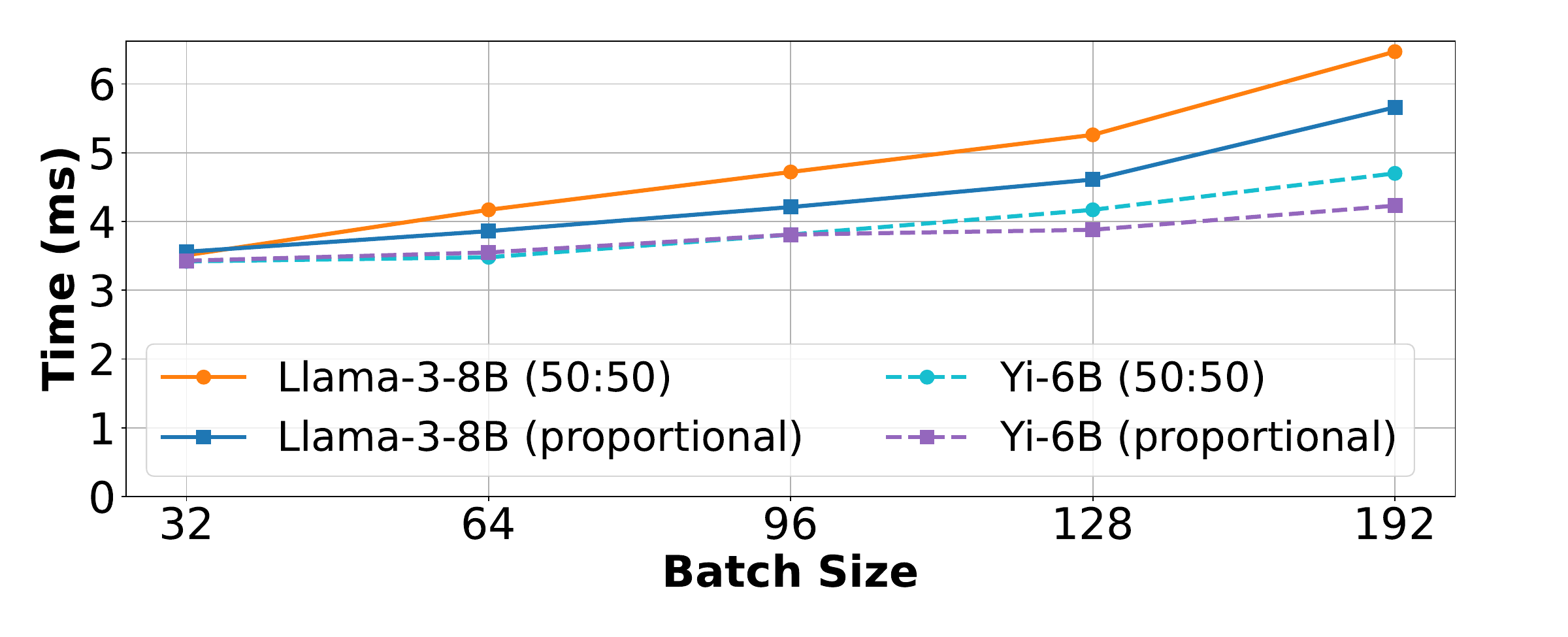}
    \caption{Effect of scheduling policy in \sysname.}
    \label{fig:eval:sens_sched}
\end{figure}

We explore two CTA scheduling policies within an SM, namely 50:50 allocation and Proportional allocation.
In 50:50 allocation, CTAs launched on an SM alternate between prefill and decodes, i.e., the first CTA performs prefill, the next decode, and so on.
This policy is agnostic to the total number of prefill and decode CTAs in the kernel. In Proportional allocation, the CTAs pick whether to perform prefill or decode depending on the total number of CTAs in the kernel.
For example, if 50 prefill and 100 decode CTAs are required, the first CTA on each SM will perform prefill, the next two CTAs will perform decode, then repeat. \autoref{fig:eval:sens_sched} shows the latency of \sysname with these policies for 8K context length and varying decode batch sizes on \yismall and \llamagqa.
We notice that as the load increases (greater batch size), the performance of Proportional improves over 50:50 allocation.
Proportional allocation spreads out the less frequent operations allowing better operational overlap and reduced resource contention, performing up to 14\% better than a 50:50 allocation scheme.

\subsubsection{Limiting Prefill Splits} \sysname reduces attention computation time with the default FlashDecoding-style splitting along the KV dimension. However, limiting the number of splits further improves performance. For example,~\autoref{tab:eval:limitingsplits} shows that in the last four chunks of a 16K prompt, co-running with 64 decode requests of the same context length, limiting the number of splits in prefill attention computation nearly doubles the speedup of \sysname over \faserial.

\subsubsection{Sensitivity to Workload}
\label{sec:eval:ablation:varyingpd}
\begin{figure}[t!]
    \includegraphics[width=0.95\columnwidth]{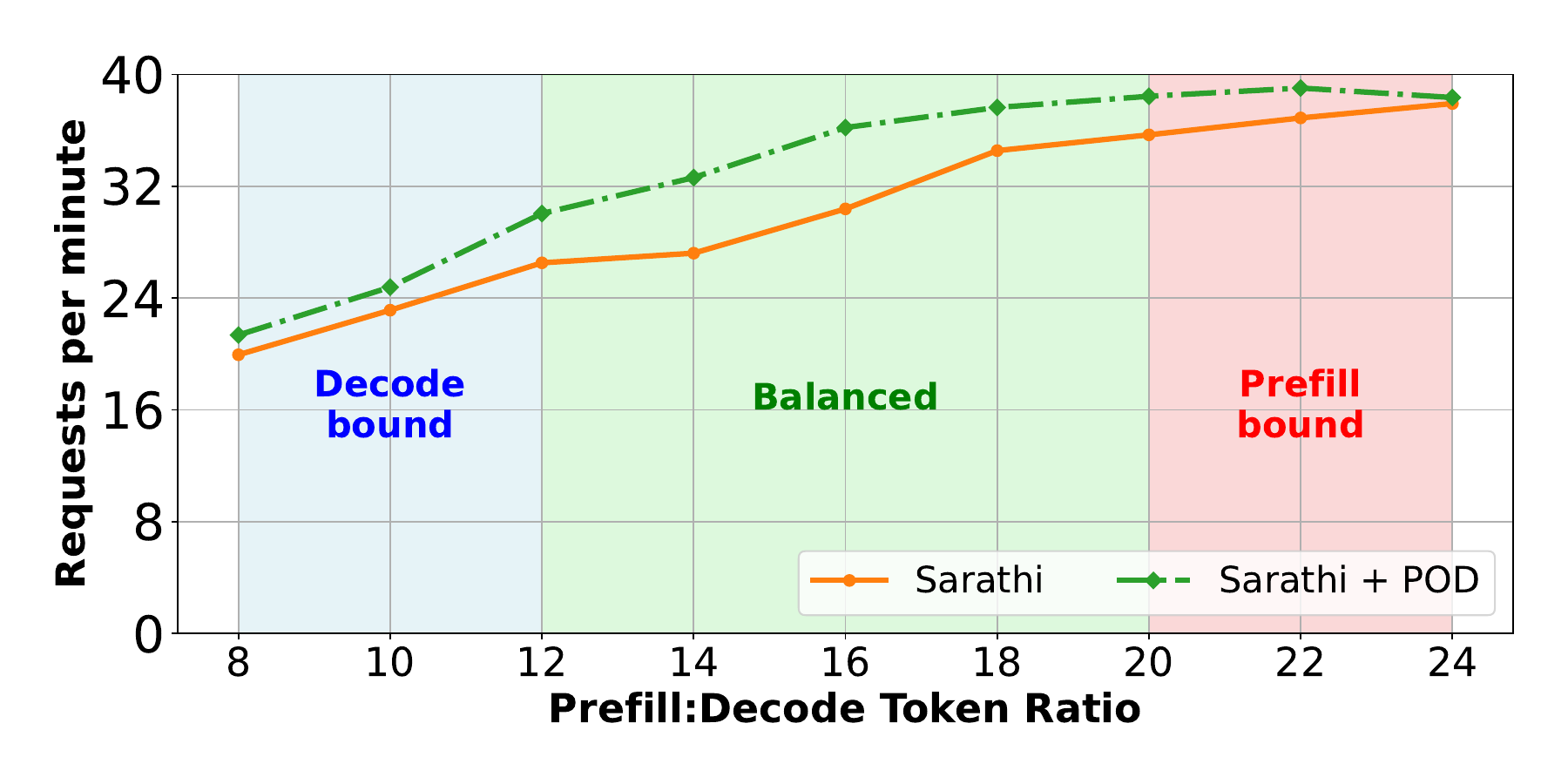}
     \caption{Request processing throughput under varying workload distribution (model: \llamagqa, TP-2). }
     \label{fig:eval:offline:varyingpd}
\end{figure}

\sysname accelerates the execution of hybrid batches and hence its impact on overall performance depends on how many iterations consist of hybrid batches in a given workload. A workload that is highly dominated by either prefills (high P:D ratio) or decodes (low P:D ratio) is likely to experience little benefit with \sysname. To understand the effect of varying P:D ratio, we benchmark \llamagqa with a total of 2048 requests, each consisting of $\approx$ 16.5K tokens, but with varying P:D ratio (in the range of 8 to 24) e.g., if the P:D is 10, then a request contains $\approx$15K prefill tokens and $\approx$1.5K decode tokens. 
\autoref{fig:eval:offline:varyingpd} shows that \sarathipod outperforms \sarathi over varying workload mixes. The peak gains occur in the P:D range of 12 to 18 because most batches are hybrid batches in this regime. In contrast, many iterations run decode-only batches when P:D ratio is lower than 12 (or prefill-only batches when P:D ratio is higher than 18).

\begin{table}[]
\scalebox{0.9}{
\begin{tabular}{c|c|c|c}
\multirow{2}{*}{Chunk Id} & \multirow{2}{*}{\faserial} & \multicolumn{2}{c}{\sysname} \\
         &  & Vanilla split & Limited split [Ours] \\ \toprule
        28 & 1.93 & 1.68 ($0.87\times$) & 1.45 ($0.75\times$)\\
        29 & 1.96 & 1.69 ($0.86\times$) & 1.45 ($0.74\times$)\\
        30 & 1.98 & 1.71 ($0.86\times$) & 1.45 ($0.73\times$)\\
        31 & 1.99 & 1.71 ($0.86\times$) & 1.46 ($0.73\times$)\\ \bottomrule
\end{tabular}}
    \caption{Per-layer attention runtime (ms) of last four prefill chunks of a prompt, co-running with decode batch size 64 (model: \llamagqa, context length: 16K, chunk size: 512).}
    \label{tab:eval:limitingsplits}
\end{table}

\section{Related Work}

Optimizing GPU execution and LLM serving systems is an active area of research~\cite{orca,vllmsosp,sarathiserve2024,vattention2024,leanattention,nanoflow2024,distserve2024,tetriinfer,splitfuse2024,patel2023splitwise,dynamollm2024,loongserve2024,mnemosyne2024,moedeployment,moe-lightening,turboattention,sglang2024,cusync2024}. 

\running{Optimizing Attention Computation}
\flashattn~\cite{flashattention} introduced the first specialized implementation of attention, fusing all its operations into a single kernel with tile-based computation. FA-2~\cite{flashattention2}  improved it further with better work partitioning and load balancing. FlashDecoding~\cite{flashdecoding} accelerates decode attention by splitting computation along the KV dimension. FlashDecoding++~\cite{flashdecoding++}  uses asynchronized softmax, double-buffered flat GEMM optimizations, and dataflow-based hardware resource adaptation to accelerate decode. LeanAttention~\cite{leanattention} follows Stream-K reduction~\cite{streamk2024} of tiled calculation to enable better load distribution across SMs for decodes. \flashinfer~\cite{flashinfer} introduced shared-prefix based optimized attention kernels. Compared to works that separately handle prefill and decode, \sysname jointly optimizes and fuses them into a single kernel.


FA-3~\cite{flash-attention-3} is a recent addition to the \flashattn family of kernels. It leverages new features available in the NVIDIA Hopper architecture, exploiting the asynchrony of Tensor Cores, the Tensor Memory Accelerator, and the Special Function Units. FA-3 was under active development at the time of writing this paper and hence we leave extending \sysname support to FA-3 and Hopper architecture for future work.


\running{Operation Fusion}
Kernel fusion is a commonly used technique for improving GPU performance. Elastic kernels~\cite{Elastic_kernel_ASPLOS13} proposes restricting resources to enable running multiple kernels concurrently. However, this method provides no guarantee of intra-SM co-location. To overcome this, ISPA~\cite{ISPA_TOC23} deploys a predetermined number of CTAs for each kernel, less than the number of CTAs that run concurrently on the GPU. Significant a priori profiling is used to determine the appropriate CTA sizes to allow for both kernels to execute concurrently. This can be tedious for attention kernels with dynamically changing input sizes, and makes load balancing between the prefill and decode operations difficult, as one operation completing early leaves resources underutilized. HFuse~\cite{HFuse_CGO_2022} fuses operations in warp-parallel fashion, providing source-to-source compilation tools to fuse kernels. SM-centric scheduling~\cite{smcentric:ICS:2015} uses the SM counter to assign work to CTAs, which we leverage in \sysname. 

\running{Optimizing LLM Inference}
Optimizing LLM serving systems is an active area of research 
~\cite{sarathiserve2024, splitfuse2024, patel2023splitwise, sheng2023fairness, moedeployment, vllmsosp, orca, dynamollm2024, fastserve, loongserve2024, powerinfer2023}. 
Orca ~\cite{orca} introduced iteration-level scheduling to eliminate compute fragmentation when requests of different lengths are batched together. PagedAttention~\cite{vllmsosp} and vAttention~\cite{vattention2024} proposed different techniques for dynamic memory management for LLM inference. \sarathiserve~\cite{sarathiserve2024} leverages chunked prefills to enable stall-free batching. In contrast, Splitwise~\cite{patel2023splitwise}, DistServe~\cite{distserve2024} and TetriInfer~\cite{tetriinfer} disaggregate the prefill and decode phases onto different GPU nodes to avoid interference between these phases. Various recent works have also proposed overlapping compute with communication to improve resource utilization~\cite{coconetasplos, flux2024, googleoverlap}.

Similar to \sysname, NanoFlow~\cite{nanoflow2024} also targets improving intra-device resource utilization, albeit with a contrasting approach. NanoFlow divides a batch into smaller operation-level nano-batches and schedules them in a way that overlaps operations with complementary resource profiles via CUDA streams. In contrast, \sysname tries to maximize resource-utilization within a given batch by fusing prefill and decode attention computation. While NanoFlow requires large batch sizes in order to benefit from batch splitting, \sysname is useful when attention consumes a significant amount of time. Therefore, NanoFlow seems more suitable for small-context scenarios whereas \sysname targets long-context scenarios that depend on hybrid batching for efficient LLM serving.
\section{Conclusion}

We introduce \sysname{} --- the first attention kernel specialized to compute prefill and decode attention in parallel such that both compute and memory bandwidth of a GPU can be utilized simultaneously. \sysname enables efficient hybrid batching based LLM inference by accelerating attention computation by up to $59\%$ (mean 28\%) compared to using independently optimized prefill and decode attention kernels. \sysname also improves the end-to-end serving throughput by up to $22\%$, while significantly reducing latency over state-of-the-art LLM serving systems \sarathiserve and \vllm.

\begin{acks}
We thank our shepherd Tim Rogers, the anonymous ASPLOS reviewers and Ajay Nayak for their valuable feedback on various aspects of the paper. We also thank Zihao Ye for various helpful discussions on \sysname and FlashInfer. Aditya K Kamath and Simon Peter are supported by National Science Foundation grant CNS-2212580.
\end{acks}

%
%
%
%
%






\appendix
\section{Artifact Appendix}

\subsection{Abstract} 
\sysname{} is a GPU  kernel that overlaps prefill and decode attention operations for large language models.
\sysname{} is built on top of \flashattn kernels (v2.6.1) \cite{flashattention2} and is integrated with  \sarathiserve{}~\cite{sarathiserve2024} -- a state-of-the-art hybrid batching based LLM inference scheduler. 

\subsection{Artifact check-list (meta-information)}


{\small
\begin{compactitem}[\labelitemi]
  \item {\bf Compilation: } CUDA 12.4, GCC 11.4.
  \item {\bf Model: } \llamamha~\cite{llama-2-7b-hf}, \llamagqa~\cite{llama-8b-hf}, \yismall~\cite{yi-6b-200k-hf}.
  \item {\bf Data set: } \arxivsummarization~\cite{arxiv}.
  \item {\bf Run-time environment: } Ubuntu 22.04, CUDA 12.4, Python 3.12, and PyTorch 2.4.
  \item {\bf Hardware: } 1--2 NVIDIA A100 80 GB GPUs, x86 machine.
  \item {\bf How much time is needed to prepare workflow?: } 1 minute with Docker image. 1--2 hours if installing from source.
  \item {\bf How much time is needed to complete experiments (approximately)?: } Approx. 18 hours.
  \item {\bf Publicly available?: } Yes.
  \item {\bf Archived (provide DOI)?: } 10.5281/zenodo.14770841
\end{compactitem}
}

\subsection{Description}

\subsubsection{How to access}
We provide the source code in various forms: Docker container (see~\ref{appendix:sw}), GitHub repository (\url{https://github.com/microsoft/vattention/tree/main/pod_attn}), and Zenodo (\url{https://doi.org/10.5281/zenodo.14770840}).


\subsubsection{Hardware dependencies}
This artifact requires an x86 machine with 2 NVIDIA A100 GPUs with 80GB memory each. If only one GPU is available, all experiments can be conducted in full, except for \autoref{table:eval:online:arxiv} and the results for \llamamha and \llamagqa in \autoref{fig:eval:throughput:offline}.

\subsubsection{Software dependencies}\label{appendix:sw}
\sysname{} has been tested on a machine with Ubuntu 22.04.
All other software dependencies are resolved while installing.

\subsubsection{Data sets}
Some experiments are based on the \arxivsummarization dataset. We use a subset of the dataset available in the \verb|traces/| folder of the artifact. 

\subsubsection{Models}
This artifact evaluates \yismall, \llamamha and \llamagqa. Accessing \yismall and \llamamha is straightforward but accessing \llamagqa requires logging into huggingface with the user's private token (\verb|HF_TOKEN| below):
\begin{Verbatim}[frame=single,rulecolor=\color{black},fontfamily=zi4,fontsize=\small]
$ huggingface-cli login --token HF_TOKEN
\end{Verbatim}

\subsection{Installation}

We provide two methods of installing and testing: using Docker (recommended) or manual installation.

\subsubsection{Docker installation (recommended)}
We provide a docker image for \sysname with all its dependencies pre-installed. 
You can launch the docker container and navigate to the artifact directory as follows:
\begin{Verbatim}[frame=single,rulecolor=\color{black},fontfamily=zi4,fontsize=\small]
$ docker run --gpus all -it \
  -p 8181:8181 --rm --ipc=host --cap-add=SYS_ADMIN \
  rnp1910/pod_attention:asplos_25_pytorch_run
$ cd /workspace/vattention/pod_attn  
\end{Verbatim}

\subsubsection{Manual installation}
For manual installation, we can download POD-Attention (available in \href{https://github.com/microsoft/vattention}{vAttention} repository) to home directory to install it. 
We use Anaconda for the appropriate versions of CUDA, Python, and PyTorch. 
This can take up to 2 hours.
\begin{footnotesize}
\begin{lstlisting}[frame=single,rulecolor=\color{black}]
$ git clone \
  https://github.com/microsoft/vattention.git
$ cd vattention/pod_attn/
# Install miniconda; skip if already installed
$ make install_miniconda
$ bash # Refresh shell and activate
$ conda activate pod_attn
# Install CUDA Toolkit
(pod_attn)$ conda install -y -c \
  conda-forge cuda-toolkit=12.4.0
# Install dependencies
(pod_attn)$ pip install -r requirements.txt
(pod_attn)$ pip install flashinfer==0.1.5 \
  -i https://flashinfer.ai/whl/cu124/torch2.4
# Install POD-Attention and vAttention
(pod_attn)$ make install_all
\end{lstlisting}
\end{footnotesize}
\ignore{
The following installs Anaconda in the home directory: 
\begin{footnotesize}
\begin{lstlisting}[frame=single,rulecolor=\color{black}]
# Create conda directory
mkdir -p ~/miniconda3
# Download conda to home directory
wget https://repo.anaconda.com/miniconda/Miniconda3-latest-Linux-x86_64.sh -O ~/miniconda3/miniconda.sh
# Install conda
bash ~/miniconda3/miniconda.sh -b -u -p ~/miniconda3
# Init
~/miniconda3/condabin/conda init
# Refresh shell to activate
bash
\end{lstlisting}
\end{footnotesize}
}

\subsection{Experiment workflow}
The source code for \sysname{} kernel is available in the \verb|vattention/pod_attn/| folder. 
Our evaluation primarily contains two kinds of experiments: attention performance (Figures 1, 6, 10, 11, 13, 14) and end-to-end LLM performance (\autoref{fig:eval:throughput:offline} and \autoref{table:eval:online:arxiv}). \autoref{fig:design:parallelism} evaluates various kernel fusion strategies with a micro-benchmark. Most of these require only one GPU except for \autoref{table:eval:online:arxiv} and \autoref{fig:eval:throughput:offline} (for \llamamha and \llamagqa) that require two GPUs. Use the Makefile present in the \verb|vattention/pod_attn/| folder to run experiments as follows:

\begin{footnotesize}
\begin{lstlisting}[frame=single,rulecolor=\color{black}]
make figure1  # 2 minutes; sudo used by script
make figure6  # 2 minutes
make figure7  # 2 minutes
make figure10 # 1 minute; sudo used by script
make figure11 # 2 hours
make figure12 # 9 hours
make figure13 # 1 minute
make figure14 # 1 minute
make table6 # 4 hours
\end{lstlisting}
\end{footnotesize}

\subsection{Evaluation and expected results}
The artifact scripts redirect the raw output numbers and logs to \verb|output/| folder, while the plotted graphs can be found in the \verb|graphs/| folder. Tables are saved as CSVs in the same folder. Results may have minor runtime variations from those reported in in the paper, but general trends should hold.









\bibliographystyle{ACM-Reference-Format}
\balance
\bibliography{main}

\end{document}